\documentclass[a4paper,fleqn]{cas-dc}

\usepackage{amsmath,amssymb}
\usepackage{lmodern} 

\usepackage[numbers]{natbib}

\makeatletter
\AtEndDocument{\immediate\write\@auxout{\string\csxdef{lastpage}{\the\numexpr\value{page}-1\relax}}}
\makeatother

\def\tsc#1{\csdef{#1}{\textsc{\lowercase{#1}}\xspace}}
\tsc{WGM}
\tsc{QE}

\begin{document}
\let\WriteBookmarks\relax
\def\floatpagepagefraction{1}
\def\textpagefraction{.001}

\shorttitle{Amortized Set Prediction for Inverse IFS Reconstruction }    

\shortauthors{Y. Yamaguti}  

\title [mode = title]{Amortized Set Prediction for Inverse IFS Reconstruction from Density Maps}  



%

\author[1]{Yutaka Yamaguti}[orcid=0000-0002-1739-3387]

\cormark[1]

\ead{y-yamaguchi@fit.ac.jp}


\credit{Conceptualization, Methodology, Software, Validation, Formal analysis,
Investigation, Data curation, Writing -- original draft, Writing -- review \&
editing, Visualization}

\affiliation[1]{organization={Faculty of Information Engineering, Fukuoka Institute of Technology},
            addressline={3-30-1 Wajiro-higashi, Higashi-ku},
            city={Fukuoka},
            postcode={811-0295}, 
            state={Fukuoka},
            country={Japan}}

\cortext[1]{Corresponding author}



\begin{abstract}
Iterated Function Systems (IFS) generate self-similar fractals from a few
contractive affine maps. 
The forward map from parameters to images is inexpensive and well understood, whereas the inverse problem is difficult and is typically solved by per-image optimization.
We replace this loop with a single forward pass of a learned estimator that predicts the affine-map set directly from a visit-frequency density map, thereby amortizing the inverse problem. 
The design follows two constraints. 
First, density maps do not uniquely identify IFS parameters, so we evaluate by reconstruction rather than parameter recovery and handle unordered map sets by Hungarian matching. Training nevertheless matches the ground-truth parameters: reconstruction-only training gives a better one-shot output, but refinement from its predictions converges to poorer minima, whereas refinement from parameter-matched predictions keeps improving.
Second, the fully known forward model lets us generate exact synthetic training pairs and supports image-only test-time refinement.
On in-distribution tests, amortized initialization plus a few refinement steps gives a better quality--speed trade-off than equal-budget random-start optimization, and over $1000$ steps it reaches high-quality reconstructions more often. 
It is robust to the sampling settings of the observation, and on an external benchmark outside its training prior it matches the mean accuracy of a recent hour-long optimizer within a minute. 
On real images (MNIST and Fashion-MNIST), it improves density metrics on average over a published per-image optimizer while being roughly $2$ to $500$ times faster per image.
\end{abstract}





\maketitle

\section{Introduction}
\label{sec:intro}

An Iterated Function System (IFS) generates a self-similar fractal as the
attractor of a small set of contractive affine maps~\cite{hutchinson1981fractals,barnsley1988fractals}.
The forward direction, which renders an image from a given set of maps, is well
understood and inexpensive to compute. The inverse direction, which recovers a
set of maps that reproduces a given image, is intrinsically difficult. From the
classical collage method~\cite{barnsley1988fractals}, moment matching, and
chaotic optimization~\cite{vrscay_moment_inverse,mantica_sloan1989} to recent
gradient descent through differentiable
renderers~\cite{tu2023learning,djeacoumar2025fractals}, most existing approaches
to recovering a global IFS solve the inverse problem separately for each image.
The few learned single-pass estimators either serve only as an initial
population for a subsequent search~\cite{graham2021nnifs} or estimate
representations other than a global affine IFS: Julia-set and L-system
parameters~\cite{liu2024inferring} and partitioned, image-domain fractal codes
for compression~\cite{poli2022selfsimilarity} (Section~\ref{sec:related}). Recent
methods based on differentiable rendering, in particular, run
hundreds to thousands of gradient steps per image over an objective landscape
that is non-convex, multimodal, and prone to poor local
minima~\cite{mantica_sloan1989,tu2023learning,djeacoumar2025fractals}.

We instead amortize the inverse problem. Given a visit-frequency density map, a learned estimator predicts the $n$ affine maps in a single feed-forward evaluation. The prediction can be used directly or as an initialization for subsequent per-image refinement. The design is shaped by two facts: the inverse problem is ill-posed, and the forward model is inexpensive and fully known.

The problem is ill-posed. Beyond the trivial permutation ambiguity of an unordered
map set, it is genuinely non-identifiable: nearly identical density maps can arise
from very different parameter sets that are not reorderings of one another, and
across our test distribution the distance between density maps is essentially
uncorrelated with the distance between parameter sets
(Section~\ref{sec:identifiability}). 
Success therefore cannot be defined as recovery of the true parameters. 

Two design choices follow. First, we compare map sets on the quotient by
permutations through a Hungarian set loss, so the estimator is never penalized for
merely reordering its maps. 
Second, we exploit the known renderer: the generative process is inexpensive to sample and provides exact labels, so we train on self-generated parameter--density pairs without external annotation.
We nevertheless train by matching the ground-truth parameters rather than the reconstruction itself. Reconstruction-only training is feasible and even gives a better one-shot output, but refinement from its predictions makes little progress and converges to poorer minima of the reconstruction error, even when the predictions are perturbed. The ground-truth parameters reproduce the observation up to the random fluctuation of rendering (the reconstruction noise floor). From predictions trained to match them, refinement reaches better minima and keeps improving.
The same forward model drives an optional test-time refinement: a few gradient steps on a reconstruction distance computed from the observed image alone, without any ground truth.

As a consequence, the learned estimator is useful not only as a fast one-shot predictor but also as an initializer for per-instance optimization.
On in-distribution data, a few refinement steps from the amortized output outperform random-start per-instance optimization under the same objective and compute budget, and remain better than a random-start optimizer given twice the time. Longer optimization shows that the advantage is not only computational: the amortized initialization reaches high-quality solutions more reliably.

The advantage is not universal. It persists under structure-preserving distribution shifts, reflections, and changes in how the observation is rendered, but weakens when the test attractors have genuinely novel structure. On the external benchmark of Djeacoumar et al.~\cite{djeacoumar2025fractals}, about a minute of refinement per image (a few seconds in batches) matches the mean accuracy of their hour-long per-image search, although that search is more accurate on most individual fractals. On $16$ fractals from our own generator, their official code with its default settings falls behind ours. On MNIST and Fashion-MNIST, compared head to head with the per-image optimizer of Tu et al.~\cite{tu2023learning}, the method is faster and more accurate under density metrics, whereas saturated-occupancy metrics favor the per-image optimizer. These claims are quantified in Section~\ref{sec:experiments}.

Our contributions are threefold.
\begin{enumerate}
  \item \textbf{A reconstruction-oriented formulation of inverse IFS estimation.} We show that density maps do not uniquely determine the underlying affine-map set, and formulate prediction as permutation-invariant set estimation with reconstruction as the evaluation criterion; on the same backbone, set estimation clearly outperforms regression in a canonical order~\cite{graham2021nnifs}. Ground-truth parameters are used only as a training signal, and we show that they are preferable to reconstruction in that role: refinement keeps improving parameter-matched predictions, whereas it barely improves predictions trained on reconstruction alone.

  \item \textbf{A better quality--speed frontier from amortized initialization.} 
  On in-distribution data, light refinement from the amortized prediction outperforms random-start optimization across all reconstruction metrics at equal budget, and remains better than a doubled-budget random baseline. Extending refinement to $1000$ steps shows that this gap reflects more reliable convergence, not only faster early progress.

\item \textbf{Direct comparison with prior per-image optimization.} 
On the benchmark of Djeacoumar et al.~\cite{djeacoumar2025fractals}, refinement from our prediction matches the mean accuracy of their hour-long search in about a minute per image and fails less often, while their search is better on most individual fractals; on $16$ of our own test fractals, their official code with its default settings falls behind (Section~\ref{subsec:external}).
We also compare directly with Tu et al.~\cite{tu2023learning} on identical targets, under a common rendering condition and two families of metrics, and obtain a two-sided result:
our method gives lower errors under density criteria, while per-image optimization
is favored under saturated occupancy. Our method is roughly $2$ to $500$ times
faster for a single image ($12$ to $2600$ times in batches), and its output also serves as an effective initialization for Tu's optimizer
(Section~\ref{subsec:mnist_tu}).
\end{enumerate}

\section{Related Work}
\label{sec:related}

Prior neural approaches to IFS inference have not combined three ingredients that are central here: permutation-invariant prediction of affine map sets, reconstruction-based evaluation through the known renderer, and amortized initialization followed by light per-image refinement. We review the relevant work in three areas: inverse IFS methods, set prediction, and differentiable rendering with amortized inference.

\subsection{The inverse IFS problem}
IFS were formalized by Hutchinson~\cite{hutchinson1981fractals} and
Barnsley~\cite{barnsley1988fractals}: a finite set of contraction maps determines
a unique attractor. The inverse problem of finding an IFS whose attractor
resembles a given image can be approached through the Collage
Theorem~\cite{barnsley1988fractals}; making the union of the images of the maps
(the collage) close to the target also brings the attractor close to it.
Classically, this inverse problem has been tackled by moment
matching~\cite{vrscay_moment_inverse}, combined wavelet and moment
methods~\cite{rinaldo_zakhor1994}, and evolutionary
computation~\cite{ea_inverse_ifs}. Jacquin's fractal image
compression~\cite{jacquin1992fractal} turned it into a practical partitioned-IFS
scheme that partitions the image and fits local contraction maps. All of these
try to solve an optimization problem separately for each image.

Tu et al.~\cite{tu2023learning} introduced gradient-based fitting of IFS parameters through a differentiable renderer. 
Their method differentiably renders chaos-game samples, splats the resulting point cloud with RBF kernels, clamps pixel values to $[0,1]$, and minimizes pixel MSE to the target. 
Because dense regions saturate after clamping, this objective measures attractor
occupancy more than visit-frequency density, whereas our objective targets the
density itself.
Our comparison in Section~\ref{subsec:mnist_tu} accordingly reports both Tu's
occupancy metric and our density metrics, so that neither method is evaluated
solely under the other's objective.
Djeacoumar et al.~\cite{djeacoumar2025fractals} render chaos-game points by
differentiable Gaussian splatting at high resolution, parameterize each map so that
it is contractive by construction, and interleave gradient descent with
simulated-annealing candidate search to escape poor local minima, at a cost of about
an hour of optimization per image.
Both methods remain per-image optimizers; we replace this loop with an amortized predictor and use its output as an initialization for optional refinement.
We compare with Djeacoumar et al.\ on their own benchmark (Section~\ref{subsec:external}) and re-implement their annealing search in our objective (Section~\ref{subsec:convergence}).

Feed-forward neural estimators for this inverse problem have also appeared. Graham
and Demers~\cite{graham2021nnifs} regress the $6n$ affine parameters directly from
a binary fractal image with a residual CNN, canonically ordering the maps and
training on parameter MSE. As they report, the predicted parameters can be close
while the reconstructed attractor looks markedly different, so the outputs serve
only as an initial population for a search algorithm; a reconstruction
(Hausdorff) loss was left aside because of its cost and the lack of a contractivity
guarantee. 
In contrast, we use the forward model so that reconstruction serves as the
criterion for evaluation and refinement, while retaining parameter matching
as the training signal; on the same backbone we compare it with canonically
ordered regression and with reconstruction-based training
(Section~\ref{subsec:training_signal}).
Liu et
al.~\cite{liu2024inferring} infer Julia-set and L-system parameters with a
multi-head autoencoder, using an image-reconstruction decoder as a semi-supervised
regularizer over unlabeled images. Their targets are Julia sets (two scalars) and
L-systems (grammar strings) rather than a set of affine contraction maps, so no
permutation structure arises, and their reconstruction acts as a regularizer
against overfitting rather than as the success criterion or a refinement objective.
In contrast, the present formulation predicts unordered affine-map sets and evaluates success by reconstruction.

Amortization of fractal encoding has also been studied for image-domain
representations. In the Neural Collages of Poli et al.~\cite{poli2022selfsimilarity},
a hypernetwork outputs, in one forward pass, the coefficients of a collage operator, a
soft generalization of the partitioned IFS of fractal image
compression~\cite{jacquin1992fractal}, and is trained by the reconstruction error of the
operator's fixed point. Its applications include compression, generative modeling, and
fractal art. The quantities being estimated differ from ours. A collage operator acts
linearly on pixel intensities over a fixed tiling: which source patch is shrunk onto
which target patch is fixed by the tiling, the learned coefficients are indexed by tile
position, and the fixed point is available in closed form. Hence neither the permutation
symmetry nor the non-identifiability of an affine-map set arises there. We estimate the
geometry itself, a small unordered set of planar affine maps whose attractor and
invariant measure reproduce the observed density, for which the reconstruction landscape
is non-convex. Transferred to this setting, their training principle,
reconstruction error without ground truth, yields a better one-shot output than
parameter matching but a poorer starting point for refinement. Amortized training on
the collage error of the Collage Theorem, which needs no rendering, is worse on both
counts (Section~\ref{subsec:training_signal}).

\subsection{Set prediction}
For problems whose output is an unordered set, Carion et
al.~\cite{carion2020detr} proposed the detection transformer (DETR), which emits
a fixed-size set of predictions in one shot from learnable queries, trained with
a set loss based on Hungarian matching~\cite{kuhn1955hungarian}. What we adopt
from this line is the training principle rather than the architecture: comparing
the predicted and target sets on the quotient by permutations, through an
optimal-assignment (Hungarian) set loss, is exactly what the order invariance of
an IFS map set requires (Section~\ref{subsec:setpred}). The architectural
machinery of DETR, such as learnable queries and attention decoding for large
sets of variable size, is not needed here: the number of maps is small and fixed
per model, and a convolutional encoder with an MLP head suffices
(Section~\ref{subsec:arch}).

\subsection{Differentiable rendering and amortized inference}
Differentiable splatting~\cite{yifan2019dss} propagates gradients from an image
loss to parameters through point-cloud rendering, and underlies our reconstruction
loss and refinement. Such a renderer is a tool for per-image gradient
optimization, however, and the cost of repeating that optimization for every input
remains. Amortized, or simulation-based, inference~\cite{cranmer2020frontier}
folds this repetition into a
single training phase by exploiting the cheapness of the forward model: when the
forward simulator is fully known, one self-generates (parameter, observation)
pairs, trains an estimator once, and answers a new observation with a single
forward pass. Amortized inference of a small set of primitives from an image has
been studied in 3D shape analysis~\cite{tulsiani2017primitives},
sharing the structure of fitting an estimator once rather than solving an
optimization per image. This shift from per-input optimization to one-shot
inference by a trained feed-forward network recurs throughout the history of deep
learning: neural style transfer moved from per-image
optimization~\cite{gatys2016style} to feed-forward
synthesis~\cite{johnson2016perceptual,ulyanov2016texture}. 
We bring this pattern to the inverse IFS problem and, beyond simple amortization,
combine it with optimization by using the trained output as an initializer for
per-image refinement.

\subsection{Positioning of this work}
This paper combines elements of these lines while differing from each in a specific way.
 The optimization-based lineage of fractal inverse methods~\cite{ea_inverse_ifs,tu2023learning,djeacoumar2025fractals,jacquin1992fractal}
performs per-image optimization; the feed-forward
estimators~\cite{graham2021nnifs,liu2024inferring} do not treat the maps as an
unordered set, use reconstruction at most as a training regularizer, and add no
refinement; and the amortized collage encoder~\cite{poli2022selfsimilarity} predicts
image-domain codes on a fixed tiling rather than a global IFS. We amortize the inverse problem and
apply the set-prediction framework~\cite{carion2020detr} to the unordered set of
IFS maps. 
We use amortized inference not only as a replacement for per-image optimization but also as an initializer for it, and we quantify the resulting quality--speed trade-off.

\section{Problem Formulation}
\label{sec:problem}

This section formalizes the IFS and its density-map generation (the forward
problem) and states the inverse problem we solve. We write a point in the plane
as a row vector $x\in\mathbb{R}^{1\times2}$ and an affine map as $x\mapsto xW+b$.

\subsection{Iterated function systems and attractors}
\label{subsec:ifs}
A set of $n$ contractive affine maps $\mathcal F=\{f_i\}_{i=1}^{n}$, with
$W_i\in\mathbb{R}^{2\times2}$ and $b_i\in\mathbb{R}^{1\times2}$,
\begin{equation}
  f_i(x)=xW_i+b_i,
  \label{eq:affine_map}
\end{equation}
is an iterated function system (IFS). When each $f_i$ is a contraction with
Lipschitz constant $\mathrm{Lip}(f_i)=\lVert W_i\rVert_2=s_{\max}(W_i)<1$, where
$s_{\max}$ is the largest singular value, the Hutchinson operator on compact sets
\begin{equation}
  \mathcal H(S)=\bigcup_{i=1}^{n} f_i(S)
  \label{eq:hutchinson}
\end{equation}
is a contraction in the Hausdorff metric and has a unique fixed set
$A=\mathcal H(A)=\bigcup_i f_i(A)$~\cite{hutchinson1981fractals,barnsley1988fractals}.
This set $A$ is the attractor of the IFS. The attractor is determined by the set
of maps alone and does not depend on the ordering of $\{f_i\}$; this order
invariance is the starting point for our set-prediction formulation
(Section~\ref{subsec:setpred}).

\subsection{Selection probabilities and the invariant measure}
\label{subsec:measure}
Generating a density map requires specifying the visit frequency of points on
the attractor. We assign each map a selection probability $p_i>0$
($\sum_i p_i=1$) and generate a trajectory by the stochastic iteration
(chaos game) $x_{t+1}=f_{\xi_t}(x_t)$, where $\xi_t$ is drawn independently with
$\Pr[\xi_t=i]=p_i$. Under contraction, the corresponding Markov operator
\begin{equation}
  (\mathcal M\,\nu)(\cdot)=\sum_{i=1}^{n} p_i\,\bigl(f_{i\#}\,\nu\bigr)(\cdot),
\end{equation}
where $f_{i\#}$ is the pushforward, has a unique invariant probability measure
$\mu=\mathcal M\mu$ whose support is $A$. The trajectory almost surely fills $A$
according to $\mu$~\cite{elton1987ergodic}, so a density map is a finite-sample
approximation of $\mu$.

We fix the selection probabilities proportional to the determinant of the linear
part:
\begin{equation}
  p_i=\frac{|\det W_i|}{\sum_{k=1}^{n}|\det W_k|}.
  \label{eq:prob_det}
\end{equation}
Here $|\det W_i|$ is the area-contraction ratio of $f_i$, so
Eq.~\eqref{eq:prob_det} assigns each map a visit frequency proportional to the
area its image occupies.\footnote{In the implementation, $|\det W_i|$ is clamped
below at $0.02$ before normalization for numerical stability. This floor is
inactive on the generating distribution, where
$|\det W|=s_1 s_2\ge0.04$ (Section~\ref{subsec:renderer}), and can act only when
rendering arbitrary predicted parameters.} This convention is standard in fractal
pre-training~\cite{kataoka2020fractaldb,anderson2022improving}, and the chaos game
of Tu et al.~\cite{tu2023learning} adopts the same rule. 
The key consequence is that the selection
probability becomes a deterministic function of the geometric parameters
$(W_i,b_i)$: the estimation target is the set of map parameters
$\theta=\{(W_i,b_i)\}_{i=1}^{n}$ alone, and the probabilities are not estimated
separately. A general IFS whose probabilities are
independent of the determinant is outside the scope of this setting.

\subsection{Forward problem: density-map generation}
\label{subsec:forward}
We denote the map from a parameter set $\theta$ to a density map (the renderer)
by $R$. Concretely, we generate chaos-game trajectory points with the
probabilities of Eq.~\eqref{eq:prob_det}, accumulate them into a histogram over a
$128\times128$ grid of the fixed domain $\Omega=[-1.5,1.5]^2$ (trajectory points
falling outside $\Omega$ are discarded), smooth it, and
normalize to unit sum to obtain a density map
$y=R(\theta)\in\mathbb{R}_{\ge0}^{128\times128}$ (rendering details in
Section~\ref{subsec:renderer}). The domain $\Omega$ and the resolution are fixed
across all samples and the densities are normalized to unit mass, so density maps
can be compared directly as probability-mass distributions on a common coordinate
system and scale.

This histogram renderer $R$ involves a discrete nearest-bin assignment and is
therefore not differentiable with respect to point positions. Whenever the
reconstruction objective (Section~\ref{subsec:objective}) is optimized by
gradients, namely at test-time refinement and in the training-time reconstruction
auxiliary, we use a differentiable surrogate renderer that overlays a bilinear
(triangular) kernel on the same binning convention (Section~\ref{subsec:renderer}).

The forward problem is fully known: sampling $\theta$ and passing it
through the renderer generates arbitrarily many exactly labeled $(\theta,y)$
pairs. We exploit this property for both self-generation
of training data and test-time refinement.

\subsection{Inverse problem: formulation as set prediction}
\label{subsec:setpred}
We solve the inverse of the forward problem: estimating the parameter set
$\theta$ from a density map $y$. Classical inverse IFS solves an optimization for
each input image; we replace this with a single
forward pass of a learned estimator $f_\Theta$, $\hat\theta=f_\Theta(y)$
(amortization).

The estimation target is an unordered set of maps. Since the attractor is
invariant to permutations of $\{f_i\}$, the comparison
between the estimator output $\hat\theta=\{(\widehat W_i,\widehat b_i)\}_{i=1}^{n}$
and the ground truth $\theta$ must be made on the quotient by the symmetric group
$S_n$: any evaluation should be invariant under a permutation $\sigma\in S_n$, and
we remove this ordering ambiguity by optimal assignment (Hungarian matching)
(Section~\ref{subsec:training}). The number of maps $n$ is fixed per model:
$n{=}4$ in the main experiments and $n{=}10$ in the real-image experiments
(Section~\ref{subsec:mnist_tu}). A variable number of maps within a single model
is outside the scope of this work.

\subsection{Reconstruction as the success criterion}
\label{subsec:objective}
How we measure the success of the inverse problem is central to this work. The
natural criterion is parameter recovery $\hat\theta\approx\theta$, but this is
ill-posed: the true parameter set cannot be uniquely recovered from a density
map, and nearly identical density maps can arise from very different parameter
sets that are not related by any permutation (quantified in
Section~\ref{sec:identifiability}). We therefore place the success criterion on
the well-posed notion of reconstruction:
\begin{equation}
  \min_{\hat\theta}\ \mathcal D\bigl(R(\hat\theta),\,y\bigr),
  \label{eq:objective}
\end{equation}
that is, we measure how well the density map $R(\hat\theta)$ obtained by passing
the estimated parameters through the forward problem reproduces the observation
$y$ (the reconstruction distance $\mathcal D$ is specified in
Section~\ref{subsec:setup}). At evaluation, accordingly, the parameter error
between $\hat\theta$ and $\theta$ is not a success metric but only a diagnostic.
During training, by contrast, this parameter discrepancy is what we minimize:
every self-generated pair carries a known $\theta$, which attains the
reconstruction noise floor (Section~\ref{sec:oracle}), and matching against it places the prediction where
refinement reaches better minima of the non-convex reconstruction landscape
(Section~\ref{subsec:training}). Training thus minimizes a parameter-matching
loss, while evaluation and the optional refinement use reconstruction.

\section{Method}
\label{sec:method}

This section describes the estimator, the supervised set-matching loss, the optional test-time refinement, and the renderer used for both data generation and reconstruction.

\subsection{Model architecture}
\label{subsec:arch}
The estimator $f_\Theta$ takes a density map as input and outputs, in one shot,
the parameters of $n$ affine maps $\{(W_i,b_i)\}_{i=1}^n$; it is a set
predictor. The input is the density map ($128\times128$, normalized to unit sum
and multiplied by the number of pixels $128^2$ so that the mean pixel value is
$1$) concatenated with two normalized coordinate channels $x,y$, giving three
channels. The encoder is a
wide residual CNN with channel and spatial attention. A $3\times3$ convolution
($\to48$ channels, GroupNorm, ReLU) is followed by seven residual blocks whose
output widths are $48,64,64,96,96,128,128$, where the blocks that first reach
$64$, $96$, and $128$ channels each apply stride-2 downsampling. Each block has a
residual branch of two $3\times3$ convolutions, gated by channel and spatial
attention in the style of CBAM~\cite{woo2018cbam}: the channel gate passes a
global average pool through a small MLP of reduction ratio $8$ to produce
per-channel weights (squeeze-and-excitation style~\cite{hu2018se}), and the
spatial gate concatenates the channel-wise mean and max and applies a $7\times7$
convolution to produce per-location weights. Three downsamplings reduce the spatial resolution from $128$ to
$16$, and an adaptive average pool to $8\times8$ followed by flattening gives a
$128\cdot8\cdot8=8192$-dimensional feature. A three-layer MLP
($8192\to512\to512\to n\cdot6$, with ReLU between layers) produces the raw output
$z\in\mathbb{R}^{n\times6}$.

For each map, the six raw values are split into a linear part and a translation
and bounded with $\tanh$:
\begin{equation}
\begin{aligned}
  W_i &= \tanh(z_{i,1:4})\in(-1,1)^{2\times2},\\
  b_i &= 1.5\,\tanh(z_{i,5:6})\in(-1.5,1.5)^2 .
\end{aligned}
\label{eq:affine_output}
\end{equation}
The output layer is initialized so that $(W_i,b_i)\approx(0.45\,I,\,0)$, a weak
contractive near-identity. The entrywise bound of Eq.~\eqref{eq:affine_output}
does not enforce contractivity, since a bound on the entries does not bound the
largest singular value below one, and we impose no explicit spectral constraint on
the network output (refinement adds a soft penalty above $0.75$;
Section~\ref{subsec:inference}). In practice the outputs are comfortably
contractive. Over the in-distribution test set (Section~\ref{subsec:setup}), the
largest singular value is $0.81$ for the one-shot prediction and $0.82$ after $30$
refinement steps, while the training targets satisfy $s_{\max}\le0.70$ by
construction (Section~\ref{subsec:renderer}). In every setting studied in this
paper, including the distribution shifts and the real images, the one-shot output
stays below $0.87$, and refinement up to $100$ steps keeps every map below one
(largest $0.904$ for \texttt{base} and $0.991$ for the reconstruction-only variant of
Section~\ref{subsec:training_signal}; Table~\ref{tab:contractivity}).
Longer refinement,
whose feasibility penalty is soft, occasionally pushes a map slightly past one: in the
$1000$-step runs of the training-signal comparison (Section~\ref{subsec:training_signal}), one map out of $1024$
(on the same test system for four of the five estimators) ends at $s_{\max}$ between
$1.005$ and $1.031$, and optimizing our output further under the occupancy objective of
Tu et al., which lacks our feasibility terms, gives one map out of $500$ at $1.006$ on
Fashion-MNIST (Appendix~\ref{sec:appendix_contractivity}).
Where a guarantee is required, it can be built into the output without loss of accuracy: an
estimator trained with the recipe of \texttt{base} whose predicted linear parts are
rescaled to $s_{\max}\le0.95$ performs on par with \texttt{base} (Chamfer $0.0536$
and $0.0253$ at $0$ and $30$ steps against $0.0546$ and $0.0259$;
Appendix~\ref{sec:appendix_contractivity}).

\subsection{Training signal: ground-truth parameter matching}
\label{subsec:training}
Since success is measured by reconstruction, the most direct training signal
would be the reconstruction error itself, which requires no ground truth. We
instead match the ground-truth parameters, not because reconstruction cannot be
trained, but because parameter matching gives a better starting point for the
subsequent refinement. The reconstruction landscape
through a differentiable renderer is non-convex and multimodal, with poor local
minima~\cite{mantica_sloan1989,tu2023learning,djeacoumar2025fractals}; our own
long-horizon experiment shows that refinement from uninformed initializations
frequently stalls at poor solutions (Section~\ref{subsec:convergence}). Gradient-based
training on reconstruction alone pushes each output down the local gradient of that
sample's reconstruction error, and local gradients do not tell a region that leads to a
good minimum from one that leads to a poor one. By contrast, the parameter set $\theta$ that
generated a training pair reconstructs $y$ as well as any parameter set can: its error is
the noise floor, the error that remains even between two renderings of the same
parameters (Section~\ref{sec:oracle}). Because the data are
self-generated, every training pair carries this $\theta$, and matching it pulls
the prediction toward parameter sets that reconstruct $y$ well rather than toward
whichever stationary point is nearest, so that refinement from the prediction
keeps descending. By non-identifiability (Section~\ref{sec:identifiability}),
$\theta$ is only one of many parameter sets that reproduce $y$, so the matching
loss cannot be driven to zero; $\theta$ serves as a training target, never as the
evaluation criterion. Section~\ref{subsec:training_signal} tests this reasoning:
reconstruction-only training on the same backbone gives a better one-shot output,
but refinement from it makes little progress and is overtaken by parameter
matching within $20$ steps.

Between the prediction $\{(\widehat W_i,\widehat b_i)\}_{i=1}^{n}$ and the ground
truth $\{(W_j,b_j)\}_{j=1}^{n}$, we remove the arbitrary ordering of the maps as a
quotient by optimal assignment. We first form the pairwise squared distance
\begin{equation}
  C_{ij}=\bigl\lVert \widehat W_i-W_j\bigr\rVert_{\mathrm F}^2
        +\bigl\lVert \widehat b_i-b_j\bigr\rVert_2^2,
\end{equation}
find the permutation $\sigma$ that minimizes it by the Hungarian method (linear
sum assignment), and average the matched squared distances:
\begin{equation}
  \mathcal L_{\mathrm{match}}
  =\frac1n\sum_{i=1}^{n}\Bigl(\bigl\lVert \widehat W_i-W_{\sigma(i)}\bigr\rVert_{\mathrm F}^2
  +\bigl\lVert \widehat b_i-b_{\sigma(i)}\bigr\rVert_2^2\Bigr).
\end{equation}
The optimal assignment keeps the permutation multiplicity of the map set out of
the loss.

We add a fixed-point consistency term. The fixed point $x^\ast$ of a map
$x\mapsto xW+b$ satisfies $x^\ast=x^\ast W+b$, that is, $x^\ast(I-W)=b$. We add the
fixed-point distance of the matched pairs with weight $\lambda_{\mathrm{fp}}{=}0.05$:
\begin{equation}
  \mathcal L=\mathcal L_{\mathrm{match}}
  +\lambda_{\mathrm{fp}}\,\frac1n\sum_{i=1}^{n}\bigl\lVert \widehat x^\ast_i-x^\ast_{\sigma(i)}\bigr\rVert_2^2 .
\end{equation}
The fixed point captures the location of each map directly and corrects the
coupled error of $W$ and $b$; its effect on reconstruction is small
(Section~\ref{subsec:training_signal}).

Training pairs are drawn from a self-generated rolling pool
(Section~\ref{subsec:renderer}), with batch size $128$. The optimizer is
AdamW~\cite{loshchilov2019adamw} (learning rate $10^{-3}$, weight decay $10^{-4}$)
for $50{,}000$ steps, with the learning rate decayed by a cosine schedule to
$10^{-4}$ after $30{,}000$ steps. We take the checkpoint with the best validation
loss.

\subsection{Inference and test-time refinement}
\label{subsec:inference}
Inference is a single forward pass: the trained $f_\Theta$ maps a density map to
$\{(\widehat W_i,\widehat b_i)\}$ at once (the 0-step output), which can be used as
the estimate on its own. Refinement is an optional post-processing step, applied
for any number of steps when higher reconstruction quality is needed, at the cost
of computation. Starting from the 0-step output, we run $K\in\{10,20,30,100\}$
steps of gradient-based optimization (AdamW, learning rate $5\times10^{-3}$) on
an objective computable from the observed image
alone: the density error on the differentiable renderer plus a symmetric squared
Chamfer term to a point cloud sampled from the observed image, the mean squared
nearest-neighbor distance averaged over both directions ($2\,\mathrm{CD}^2$ with
$\mathrm{CD}$ as defined in Section~\ref{subsec:setup}; weight $0.10$). Because refinement optimizes the raw affine
parameters without the output bounds of Eq.~\eqref{eq:affine_output}, the
objective also includes weak feasibility penalties: hinge terms discouraging
singular values above $0.75$, negative determinants, and translations outside the
drawing domain $\Omega$ (weights $0.1$, $0.1$, and $0.01$, respectively), which vanish
inside the corresponding bounds. The number of steps $K$ trades quality against
speed and can be chosen continuously, including $K{=}0$ (no post-processing). The
per-image optimization used as a baseline minimizes the same objective with the
same optimizer and budget from a random initialization, so the only difference
from refinement is the initialization.

\subsection{Differentiable renderer and data generation}
\label{subsec:renderer}
The attractor is obtained as the cloud of trajectory points traced by the
iterated maps, which we convert to a density map. For generation (the target) we
use a hard histogram rendering (resolution $128$, $16$ trajectories of $1024$
steps each, burn-in $128$, smoothing $\sigma{=}2$). For the training-time
reconstruction auxiliary and for refinement, we use a differentiable version
(soft splatting) that overlays a triangular (bilinear) kernel on the same binning
convention, with settings matched to the generation renderer; evaluation likewise
uses a high-fidelity renderer with the same settings, which we call the matched
renderer in Section~\ref{sec:experiments}. 
All densities are normalized to unit
sum. 
For a fixed map-index sequence, the trajectory points are differentiable functions of the affine parameters. Bilinear splatting then makes the image accumulation piecewise differentiable with respect to the point locations. During each refinement step, the sampled map-index sequence is treated as fixed in the backward pass; gradients are not propagated through the discrete sampling operation, including the dependence of the sampled indices on the probabilities. Gradients thus propagate from the image
loss to $(W_i,b_i)$, allowing the reconstruction objective (Eq.~\eqref{eq:objective})
to be optimized by gradient descent.

For data generation, each map is sampled from a fixed point
$x^\ast\sim\mathcal U(-0.75,0.75)^2$, singular values
$s_1,s_2\sim\mathcal U(0.20,0.70)$, and rotation angles
$\phi_1,\phi_2\sim\mathcal U(-\pi,\pi)$, and is constructed as
\begin{equation}
\begin{aligned}
  W&=\mathrm{Rot}(\phi_1)\,\mathrm{diag}(s_1,s_2)\,\mathrm{Rot}(\phi_2),\\
  b&=x^\ast(I-W).
\end{aligned}
\end{equation}
Not every sampled system is used: each candidate is first rendered at low
fidelity and accepted only if all trajectory points are finite, at most $20\%$
of them fall outside $\Omega$, at least $0.5\%$ of the pixels are occupied, and
no single pixel carries more than $25\%$ of the total mass. This rejection step
excludes degenerate systems that collapse to nearly a point or mostly leave the
drawing domain, so the effective training and test distribution is the
construction above conditioned on acceptance. Accepted samples are generated in
batches on the GPU, and a rolling cache of $65{,}536$ samples is regenerated with
a new seed every $1{,}000$ steps to maintain sample diversity throughout
training.

Two properties of this generating distribution are worth noting. First, since
$U{=}\mathrm{Rot}(\phi_1)$ and $V{=}\mathrm{Rot}(\phi_2)$ are rotations
(determinant $+1$) and $s_i>0$, the construction produces only
orientation-preserving maps with $\det W=s_1 s_2>0$ and excludes reflections
($\det<0$). Since the singular value decomposition of a general real $2\times2$
matrix admits orthogonal factors that may be rotations or reflections, our
generation covers only part of the representable maps. The SVD-based
parameterization we use follows the synthetic-fractal pre-training
study~\cite{anderson2022improving} and the form used by Tu et
al.~\cite{tu2023learning}; that parameterization admits reflections through a
trailing sign-flip factor $\mathrm{diag}(\pm1,\pm1)$, which we omit, restricting
the training prior to orientation-preserving maps to keep the distribution simple
and well-conditioned. The output parameterization
(Eq.~\eqref{eq:affine_output}) places no sign constraint on $W_i$, so the model
can represent $\det W<0$; the restriction is a choice of training prior, not a
limit on model capacity (reflected test systems and a prior with reflections are
examined in Section~\ref{subsec:ood}). Second, the singular value range $[0.20,0.70]$ is chosen
as follows. The largest singular value $s_{\max}{=}\max(s_1,s_2)$ is the operator
$2$-norm of the linear part (the Lipschitz constant of the map), and if every map
is contractive ($s_{\max}<1$) the Hutchinson condition guarantees a unique bounded
attractor~\cite{hutchinson1981fractals,barnsley1988fractals}. The upper bound
$0.70$ keeps the contraction strong: a weak contraction with $s_{\max}$ near $1$
mixes slowly, spreads the attractor space-fillingly so that accurate density
rendering needs long trajectories, and makes the fixed-point iteration
ill-conditioned. The lower bound $0.20$ avoids degeneracy: a map with $s_i$ near
$0$ collapses to its fixed point and contributes a point-like degenerate mass.
This range corresponds to maps that each have a spatially meaningful extent while
the whole system renders stably and quickly.

\section{Identifiability: parameter recovery is ill-posed}
\label{sec:identifiability}

This section justifies reconstruction-based evaluation by showing that the true IFS parameter set is not identifiable from a density map.
This is a property of the inverse IFS problem itself, not of our model's predictions.
We first exhibit the structural source of this non-identifiability through an analytic example, and then quantify how it appears in our test distribution.

Consider the Sierpinski triangle:
it is the attractor $A$ of an IFS of three maps $f_i(x)=\tfrac12 x+b_i$ (linear
part $W_i=\tfrac12 I$), each contracting by $1/2$ toward a vertex $c_i$. Because
$A$ is symmetric under the $120^\circ$ rotation $\mathrm{Rot}$ about its centroid
($\mathrm{Rot}(A)=A$), inserting $\mathrm{Rot}$ on the input side of each map,
$g_i=f_i\circ\mathrm{Rot}$, gives
\[
  \textstyle\bigcup_i g_i(A)=\bigcup_i f_i\bigl(\mathrm{Rot}(A)\bigr)=\bigcup_i f_i(A)=A,
\]
so $\{g_i\}$ has the same attractor. Because $\mathrm{Rot}$ maps the triangle's
vertices to one another, it permutes the maps
($\mathrm{Rot}\circ f_j=f_{\sigma(j)}\circ\mathrm{Rot}$ for a permutation
$\sigma$), and Eq.~\eqref{eq:prob_det} assigns all maps the equal probability
$\tfrac13$ since the determinants coincide. Pushing the invariance
$\mu=\sum_i\tfrac13 f_{i\#}\mu$ forward by $\mathrm{Rot}$ and applying this
identity gives
$\mathrm{Rot}_{\#}\mu=\sum_j\tfrac13 f_{j\#}(\mathrm{Rot}_{\#}\mu)$:
$\mathrm{Rot}_{\#}\mu$ is again an invariant measure, so uniqueness forces
$\mathrm{Rot}_{\#}\mu=\mu$. Consequently each $g_i=f_i\circ\mathrm{Rot}$ pushes
$\mu$ forward exactly as $f_i$ does
($g_{i\#}\mu=f_{i\#}(\mathrm{Rot}_{\#}\mu)=f_{i\#}\mu$), so $\mu$ is invariant
for $\{g_i\}$ as well (the $g_i$ share the same determinants, hence the same
probabilities), and the two systems generate exactly identical density maps. Yet the
linear part of $g_i$ is $W'_i=\tfrac12\,\mathrm{Rot}$, a rotation, differing from
the original $W_i=\tfrac12 I$. This difference cannot be absorbed by permuting the
maps, and the parameter-set distance remains large even after optimal (Hungarian)
assignment. Thus the same density map arises from parameter sets that are not
related by any reordering.

\begin{figure*}[t]
  \centering
  \includegraphics[width=0.42\linewidth]{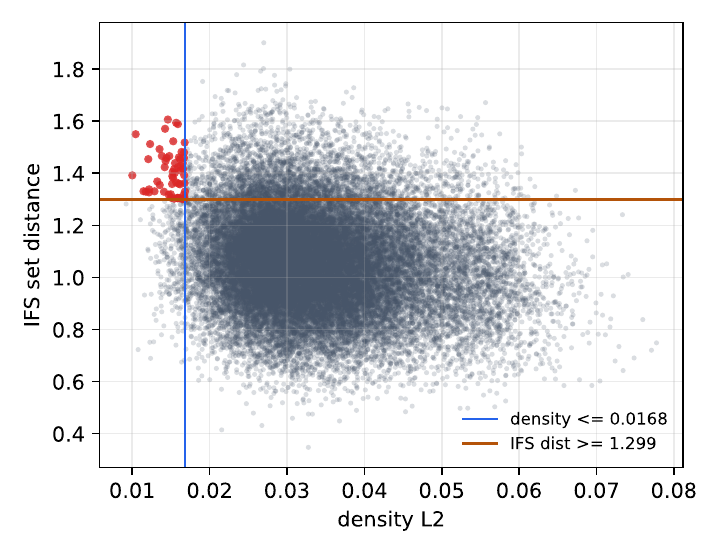}\hfill
  \includegraphics[width=0.42\linewidth]{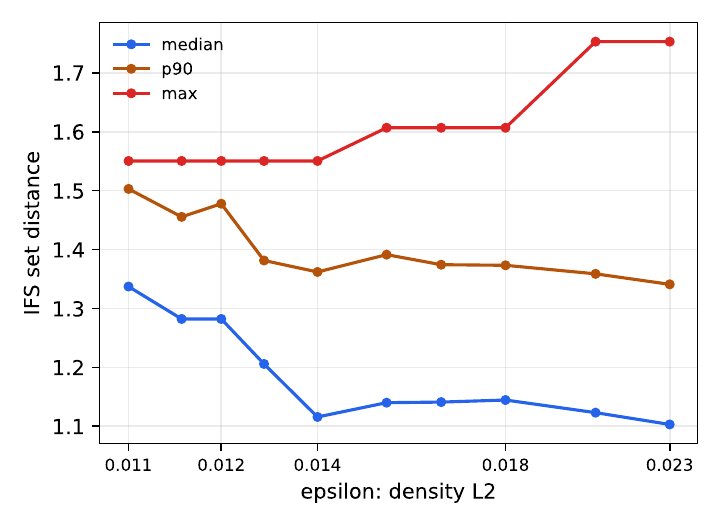}
  \caption{Identifiability (all pairs of \texttt{test256}). Left: density distance
  versus parameter-set distance (Pearson correlation $-0.146$). The $59$ near-image,
  far-parameter pairs that simultaneously satisfy the thresholds (density
  $\le0.0168$, the bottom $1\%$; parameter distance $\ge1.299$, the top $10\%$)
  are shown in red. Right:
  parameter-set distance (median/p90/max) among image pairs with density distance
  $\le\varepsilon$; it does not fall off as $\varepsilon$ shrinks.}
  \label{fig:ident}
\end{figure*}

We next quantify that such non-identifiability appears in our synthetic
distribution. Over all $\binom{256}{2}=32{,}640$ pairs of the fixed test set
\texttt{test256}, we examine the relationship between the distance between density
maps and the distance between parameter sets. The density distance is the $L_2$
distance between the unit-sum $128\times128$ density maps, and the parameter-set
distance is the composite $(W,b)$ distance after matching the map sets by optimal
assignment (Hungarian; as in Section~\ref{subsec:training}, the minimum over all
$24$ permutations for $n{=}4$), so the ordering ambiguity is removed as a
quotient. The two distances are essentially uncorrelated (Pearson $-0.146$,
Spearman $-0.141$): a ``close image implies close parameters'' relationship is
largely absent (Table~\ref{tab:ident_dist}, Figure~\ref{fig:ident}). 
Low correlation over all pairs is weak evidence on its own, since it can also arise from benign nonlinear relationships. 
Stronger evidence comes from the tail of near-identical images, which we examine next.
Even taking each sample's nearest-image neighbor, the corresponding parameter distance has a
median of $1.00$, barely below the overall median of $1.06$. The points shown in
red in Figure~\ref{fig:ident} have a small density distance ($\le0.0168$, the
bottom $1\%$ in Table~\ref{tab:ident_dist}) together with a large parameter-set
distance ($\ge1.299$, the top $10\%$ of the pairwise distribution); there are
$59$ such pairs. In representative cases, a
density $L_2$ of about $0.010$--$0.016$ corresponds to a parameter-set distance of
about $1.55$--$1.61$ ($W$ Frobenius about $0.9$--$1.2$): nearly identical density
maps can arise from very different parameter sets.

\begin{table}[t]
  \centering\small
  \caption{Pairwise distance distribution on \texttt{test256} (density distance
  and parameter-set distance).}
  \label{tab:ident_dist}
  \begin{tabular}{lrrrr}
    \toprule
    Distance & Mean & Median & $1\%$ & Max \\
    \midrule
    Density $L_2$ & 0.0340 & 0.0325 & 0.0168 & 0.0778 \\
    $(W,b)$ distance & 1.0606 & 1.0559 & 0.6535 & 1.9016 \\
    \bottomrule
  \end{tabular}
\end{table}

Non-identifiability is most precisely stated as a function of a tolerance
$\varepsilon$. Even restricting to image pairs whose density distance is at most
$\varepsilon$, the parameter-set distance does not fall off
(Figure~\ref{fig:ident}, right): shrinking $\varepsilon$ to the smallest level
present within \texttt{test256} (density $L_2\approx0.013$), the median
parameter-set distance stays at about $1.2$, comparable to or above the overall
$\approx1.06$. Moreover, this $\varepsilon$ level is on the order of the
reconstruction accuracy that refinement attains (Section~\ref{sec:experiments}),
showing that the non-identifiability is not an asymptotic artifact but appears
near the accuracy attained by our refinement. The same holds at a larger scale: in a
reference set of $131{,}072$ systems drawn from the training generator, the systems
whose density lies within $0.0168$ of a test map are at a median parameter-set
distance of $1.03$ from it (median over the $101$ test samples with at least three
such neighbors), again comparable to two unrelated systems
(Appendix~\ref{sec:appendix_ambiguity}).

Consequently, in the present synthetic distribution the problem of uniquely
recovering the true parameter set from a density map is not well posed. We
therefore place the success criterion on reconstruction and use the $(W,b)$ error
as a secondary indicator of true-value recovery and to characterize failure modes
(Section~\ref{sec:experiments}).

\section{Experiments}
\label{sec:experiments}

\subsection{Experimental setup}
\label{subsec:setup}
The estimator takes a density map as input and outputs a set of $n{=}4$ affine
maps $\{(W_i,b_i)\}_{i=1}^{4}$. Following Section~\ref{sec:identifiability}, we
place the success criterion on reconstruction: we render the attractor defined by
the predicted parameters and measure its agreement with the target density map.
The primary metrics are four complementary measures of reconstruction quality
(density squared error, Chamfer distance, $95$th-percentile Hausdorff distance,
and $2$-px coverage), defined below.

We use a fixed test set \texttt{test256} ($256$ samples with a fixed generation
seed). At evaluation, the predicted attractor is rendered with a high-fidelity
renderer whose settings are identical to those of the renderer that generated the
target density map (settings in Section~\ref{subsec:renderer}). Matching the
renderer between generation and evaluation is important for the reconstruction
error to be a meaningful measure of achievement: if the settings disagree, a
mismatch occurs in which the true parameters do not minimize the
error (Section~\ref{sec:oracle}). We call this evaluation renderer the matched
renderer. The estimator processes all $256$ samples in a single forward pass.

For notation, we compare the predicted and target attractors with one
density-based metric and three point-cloud metrics. A density map is normalized to
unit sum on a $128$-resolution grid, and we write the prediction and target as
$\widehat\rho,\rho\in\mathbb{R}^{128\times128}_{\ge0}$
($\sum_p\widehat\rho_p=\sum_p\rho_p=1$). For the point-cloud metrics, each
attractor is drawn as a trajectory point cloud, and the finite points within the
drawing domain $[-1.5,1.5]^2$ form the point sets $\widehat{X},X$ (subsampled to at
most $2048$ points each to reduce cost, $|\widehat{X}|,|X|\le2048$). For a point
$u\in\mathbb{R}^2$ and a nonempty finite set $S$, the point-to-set distance is
\begin{equation}
  d(u,S)=\min_{v\in S}\lVert u-v\rVert_2,
  \label{eq:point_set_dist}
\end{equation}
and we write $\{d(u,X):u\in\widehat{X}\}$ for the nearest-neighbor distances from
$\widehat{X}$ to $X$, and $\{d(v,\widehat{X}):v\in X\}$ for the reverse. For a
finite nonnegative sequence $w_{(1)}\le\dots\le w_{(n)}$, the empirical
$q$-quantile $Q_q$ is defined, with $h=1+q(n-1)$, by
\begin{equation}
  Q_q=w_{(\lfloor h\rfloor)}+(h-\lfloor h\rfloor)\bigl(w_{(\lceil h\rceil)}-w_{(\lfloor h\rfloor)}\bigr)
\end{equation}
(linear interpolation, with $Q_1$ the maximum).

\begin{description}
  \item[Density squared error (density SSE)] The pixelwise sum of squared
    differences of the normalized density maps,
    \begin{equation}
      \mathrm{SSE}(\widehat\rho,\rho)=\sum_{p}\bigl(\widehat\rho_p-\rho_p\bigr)^2 .
    \end{equation}
    It measures agreement of the density (the intensity distribution proportional
    to visit frequency), including where the mass concentrates.
  \item[Chamfer distance] The symmetric form obtained from the mean squared
    nearest-neighbor distances in both directions,
    \begin{equation}
    \begin{aligned}
      \mathrm{CD}(\widehat{X},X)=\tfrac12\Bigl(
      &\tfrac{1}{|\widehat{X}|}\textstyle\sum_{u\in\widehat{X}} d(u,X)^2\\
      &+\tfrac{1}{|X|}\textstyle\sum_{v\in X} d(v,\widehat{X})^2
      \Bigr)^{1/2},
    \end{aligned}
    \end{equation}
    so that $\mathrm{CD}=\delta/\sqrt2$ when every nearest-neighbor distance equals $\delta$.
    It is a monotone transform of the commonly used squared form (the sum of the two
    mean squared nearest-neighbor distances), so its scale has no effect on any
    comparison.
   It primarily measures support agreement and is less directly sensitive to visit-frequency differences than density SSE, although the trajectory samples are still drawn according to the invariant measure.
  \item[$95$th-percentile Hausdorff distance (HD95)] The larger of the $95$th
    percentiles of the nearest-neighbor distance distributions in each direction,
    \begin{equation}
    \begin{aligned}
      \mathrm{HD}_{95}(\widehat{X},X)=\max\bigl(
      &Q_{0.95}(\{d(u,X)\}_{u\in\widehat{X}}),\\
      &Q_{0.95}(\{d(v,\widehat{X})\}_{v\in X})\bigr).
    \end{aligned}
    \end{equation}
    The ordinary directed Hausdorff distance ($q=1$, the maximum) is dominated by a
    single outlier, so we use this robust version that discards the top $5\%$,
    measuring large boundary deviations without oversensitivity to outliers.
  \item[$2$-px coverage] The fraction of points with a
    neighbor in the other set within a tolerance radius $\tau$, averaged over both
    directions (denoted coverage@2px),
    \begin{equation}
    \begin{aligned}
      \mathrm{cov}_\tau(\widehat{X},X)=\tfrac12\bigl(
      &\tfrac{1}{|\widehat{X}|}\textstyle\sum_{u\in\widehat{X}}\mathbf{1}[d(u,X)\le\tau]\\
      &+\tfrac{1}{|X|}\textstyle\sum_{v\in X}\mathbf{1}[d(v,\widehat{X})\le\tau]
      \bigr),
    \end{aligned}
    \end{equation}
    with $\tau=2\cdot(3/128)\approx0.0469$ (twice the pixel width $3/128$; the full
    width of the coordinate system $[-1.5,1.5]^2$ is $3$). Under a fixed tolerance,
    it measures how completely the two attractors cover each other.
\end{description}
$\mathrm{SSE},\mathrm{CD},\mathrm{HD}_{95}$ are better when smaller and
$\mathrm{cov}_\tau$ when larger. They complement one another: if the shape matches
but the density bias differs, $\mathrm{SSE}$ worsens, and if the global match is
good but part is far off, $\mathrm{HD}_{95}$ worsens.

The standard model of this paper (denoted \texttt{base}) is the bare amortized
estimator of Section~\ref{sec:method}, trained with GT-$\theta$ Hungarian matching
and no reconstruction auxiliary.

We evaluate the one-shot output of the standard model ($0$ steps) and the results
of applying the refinement of Section~\ref{subsec:inference} for $10/20/30$ steps
(with $512$ point samples each for prediction and target). The baseline is
per-image optimization from a random initialization under the same objective and
budget (random-r4, $4$ restarts; Section~\ref{subsec:inference}), corresponding to
the per-image gradient optimization of prior work (Tu et al.\ 2023 and others). We
place both on the same wall-clock time axis (seconds per sample) for a fair
quality--speed comparison.

All timing experiments were run on an NVIDIA GeForce RTX 3090 with an AMD Ryzen 9
5950X processor, using PyTorch 2.3.0 and CUDA 12.1. Timings are wall-clock seconds
per sample, measured with a monotonic host clock (\texttt{time.perf\_counter})
around each optimization block. Because every measured block returns its recovered
parameters and losses to the host, the device-to-host copies force outstanding GPU
kernels to complete before the clock is read. Random restarts are folded into the
batch dimension and thus evaluated in parallel on the GPU where memory permits, so
the reported cost is wall-clock time rather than FLOPs, which if anything favors the
multi-restart random baseline. The official code of Djeacoumar et al.\ (Section~\ref{subsec:external})
ran in its own environment (PyTorch 2.6.0, CUDA 12.4) on the same GPU and was timed
with \texttt{time.time} around each optimization.

\subsection{Evaluation design: matched renderer and reconstruction validity}
\label{sec:oracle}
Before the main comparison, we verify the validity of the evaluation design. To
use reconstruction as the objective and metric, the renderer used for optimization
and evaluation must match the generation renderer: if they disagree, the objective
is misspecified and a non-ground-truth solution with a loss lower than that of the
true parameters can arise, as we actually observe with a low-fidelity renderer. We
therefore set the evaluation and optimization renderer to the same settings as
generation.

Under this setting, the reconstruction error of the true parameters rendered by the
matched renderer drops to about $2.7\times10^{-6}$ in density SSE. Because the
matched renderer draws with a finite number of stochastic trajectories, even the
same parameters produce a fluctuation of about $2.6\times10^{-6}$ in density SSE
from one drawing to another. Since the target density map is itself one such
drawing, no parameters can reproduce the target below this fluctuation; we call
this the reconstruction noise floor. 
The true parameters reach this floor, so under the matched renderer the reconstruction error is a valid evaluation metric.
Unless stated otherwise (the renderer-mismatch study of Section~\ref{subsec:ood} and
the comparisons with external code in Sections~\ref{subsec:external}
and~\ref{subsec:mnist_tu}), all evaluation and refinement use the matched renderer.

Minimizing reconstruction does not necessarily imply parameter recovery
(Section~\ref{sec:identifiability}): indeed, refinement from the estimator's output
improves reconstruction substantially while the $(W,b)$ error barely decreases
(Section~\ref{subsec:pareto}). We therefore treat the $(W,b)$ error not as a
failure gate but as a diagnostic (Section~\ref{sec:discussion}).

\subsection{Main comparison: the quality--speed Pareto front}
\label{subsec:pareto}
\begin{figure*}[t]
  \centering
  \includegraphics[width=0.48\linewidth]{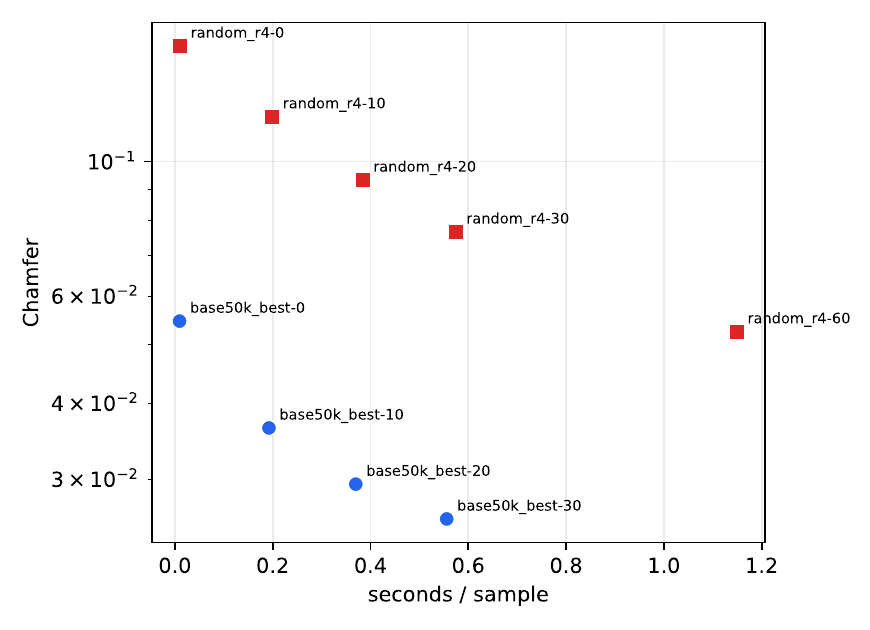}\hfill
  \includegraphics[width=0.48\linewidth]{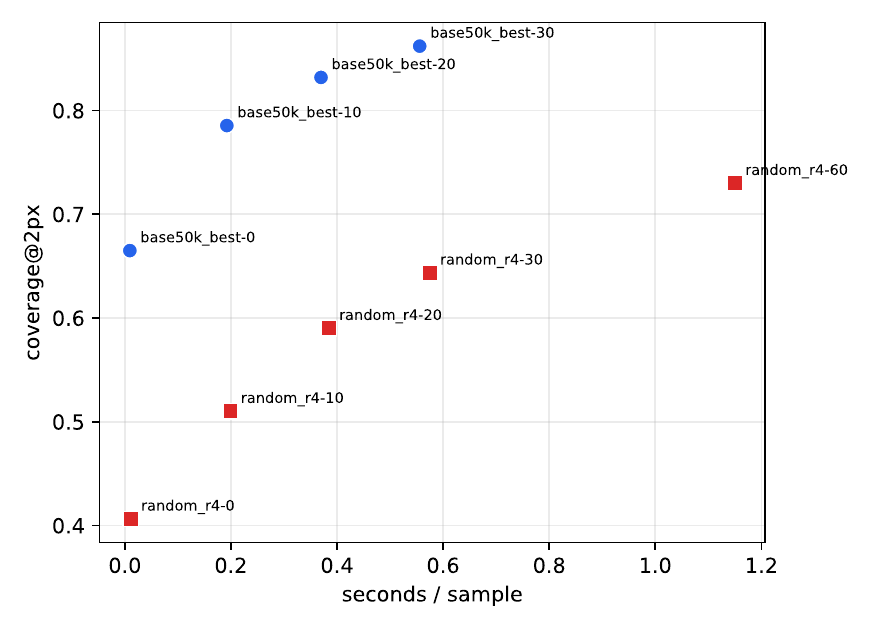}
  \caption{Quality--speed Pareto (\texttt{test256}). Horizontal axis: seconds per
  sample. Left: Chamfer (smaller is better); right: coverage@2px (larger is
  better). The \texttt{base} amortized(+refinement) curve consistently improves over
  equal-budget random-from-scratch.}
  \label{fig:hero_pareto}
\end{figure*}

\begin{figure*}[t]
  \centering
  \includegraphics[width=0.82\linewidth]{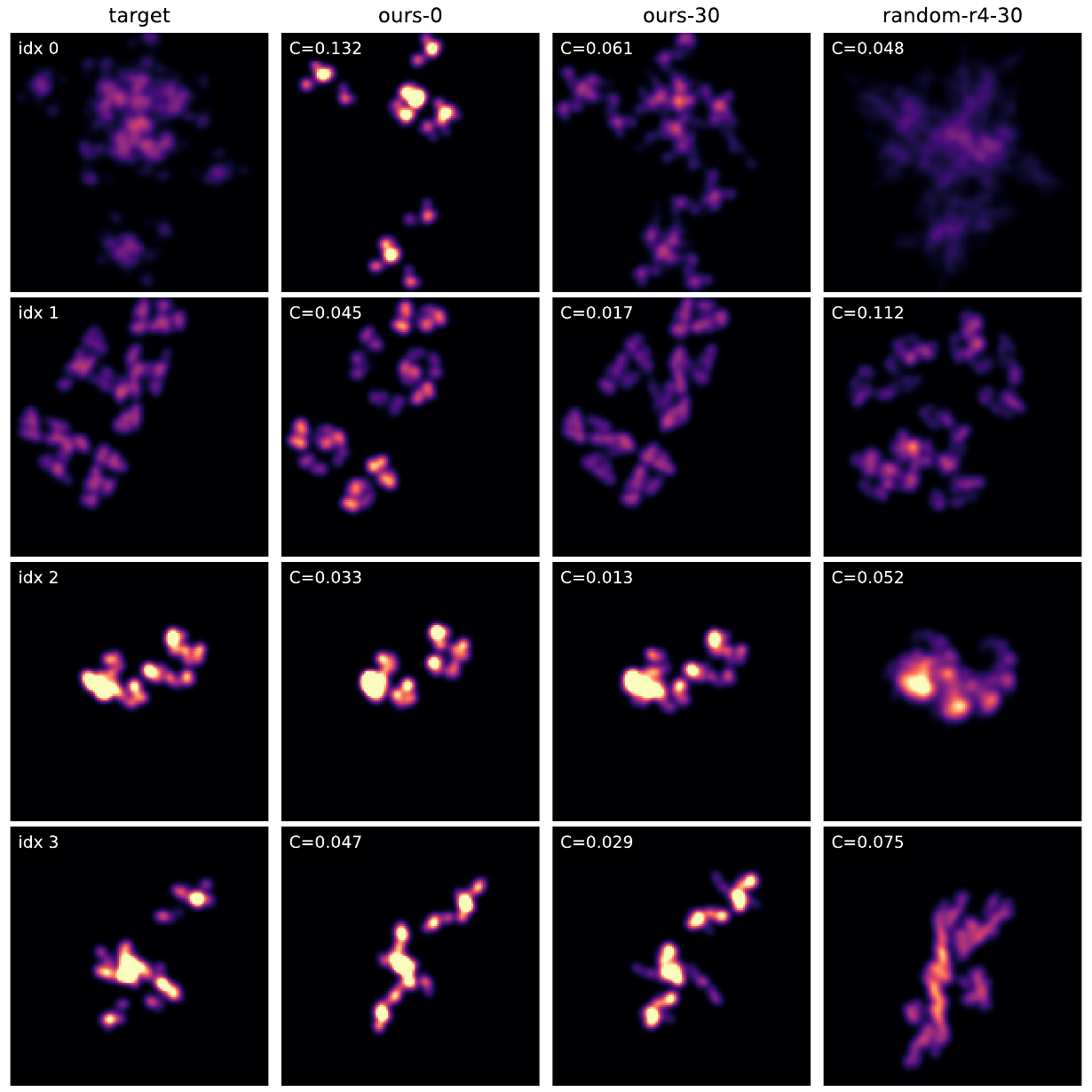}
  \caption{Qualitative examples of the main result ($4$ representative examples
  of \texttt{test256}, density maps by the matched renderer). Columns: target,
  ours-$0$ (one-shot), ours-$30$ ($+30$ refinement), and random-r4-$30$
  (equal-budget per-instance); $C$ in each cell is the Chamfer distance. At equal
  budget, ours-$30$ is closer to the target than random-r4-$30$ in most examples
  (point-cloud version in Figure~\ref{fig:hero_qualitative_points}).}
  \label{fig:hero_qualitative_density}
\end{figure*}

Table~\ref{tab:hero_pareto} and Figure~\ref{fig:hero_pareto} show the
$0/10/20/30$-step refinement of the amortized estimator (\texttt{base}) and the
equal-budget random-r4 baseline ($0/10/20/30$ steps, and $60$ steps at double the
time). Representative density maps are in Figure~\ref{fig:hero_qualitative_density},
with the corresponding point clouds in Figure~\ref{fig:hero_qualitative_points}.

\begin{table*}[t]
  \centering
  \small
  \caption{Quality--speed Pareto (\texttt{test256}). \texttt{base}
  amortized(+refinement) versus equal-budget random-from-scratch (r4). Time is
  seconds per sample; density SSE, Chamfer, and HD95 are better when smaller and
  coverage@2px when larger. The $(W,b)$ error is the Hungarian-matched
  parameter-set distance (same definition as Section~\ref{sec:identifiability}),
  reported as a diagnostic.}
  \label{tab:hero_pareto}
  \begin{tabular}{lrrrrrr}
    \toprule
    Setting & sec/sample & density SSE & Chamfer & HD95 & coverage@2px & $(W,b)$ err \\
    \midrule
    \texttt{base} 0-step      & 0.0095 & $7.85\times10^{-4}$ & 0.0546 & 0.2106 & 0.6650 & 0.405 \\
    \texttt{base} +10         & 0.192  & $3.85\times10^{-4}$ & 0.0365 & 0.1381 & 0.7855 & 0.395 \\
    \texttt{base} +20         & 0.370  & $2.76\times10^{-4}$ & 0.0295 & 0.1059 & 0.8319 & 0.394 \\
    \textbf{\texttt{base} +30} & 0.556 & $\mathbf{2.23\times10^{-4}}$ & \textbf{0.0259} & \textbf{0.0913} & \textbf{0.8621} & 0.395 \\
    \midrule
    random-r4 0-step    & 0.0110 & $1.005\times10^{-3}$ & 0.1546 & 0.5719 & 0.4063 & 0.981 \\
    random-r4 +10       & 0.199  & $7.69\times10^{-4}$ & 0.1184 & 0.4566 & 0.5102 & 0.973 \\
    random-r4 +20       & 0.384  & $6.40\times10^{-4}$ & 0.0932 & 0.3712 & 0.5901 & 0.982 \\
    random-r4 +30       & 0.575  & $5.62\times10^{-4}$ & 0.0766 & 0.3097 & 0.6434 & 0.977 \\
    random-r4 +60 ($2\times$ time) & 1.150 & $4.35\times10^{-4}$ & 0.0525 & 0.2136 & 0.7299 & 0.967 \\
    \bottomrule
  \end{tabular}
\end{table*}

The \texttt{base} refinement curve lies above the random-r4 curve on the quality--speed frontier for all four reconstruction metrics. At $0$ steps, the feed-forward prediction already improves over random-r4 at comparable time (Chamfer $0.0546$ vs $0.1546$; coverage@2px $0.665$ vs $0.406$). At about $0.56$ s/sample, \texttt{base}+30 improves over random-r4+30 on every metric and wins per sample on 239--253 of the 256 cases, depending on the metric. It also remains better than random-r4+60, despite the latter using about twice the time.

To test robustness to the training seed, we trained two additional estimators with the same recipe and evaluated all three on \texttt{test256}.
 The three-seed average of \texttt{base}+30 was density SSE $2.34\times10^{-4}\pm0.10\times10^{-4}$, Chamfer $0.0269\pm0.0009$, HD95 $0.0955\pm0.0036$, and coverage@2px $0.853\pm0.008$.

 The between-seed spread is far smaller than the gap to random-r4; for example, the Chamfer standard deviation is $0.0009$, compared with a gap of about $0.026$ to random-r4+60. Each seed remains better than random-r4 at both equal and doubled budgets, and per-sample win rates against random-r4+60 remain stable across seeds.

As the $(W,b)$-error column of Table~\ref{tab:hero_pareto} shows, refinement
improves reconstruction substantially (\texttt{base}: Chamfer $0.0546\to0.0259$)
while the $(W,b)$ error barely moves ($0.405\to0.395$); random-r4's $(W,b)$ error
also stays high, around $0.97$. Improvement in reconstruction therefore does not
imply parameter recovery (Section~\ref{sec:identifiability}), and we keep the
$(W,b)$ error as a diagnostic rather than a success criterion.

\subsection{Long-horizon refinement and convergence distribution}
\label{subsec:convergence}
The quality--speed comparison in Section~\ref{subsec:pareto} is a comparison at equal
budget, which alone cannot distinguish whether random initialization is merely
slower to converge or becomes trapped in worse local minima. Because the
reconstruction landscape of the inverse IFS problem is non-convex, multimodal, and
prone to poor local minima~\cite{mantica_sloan1989,tu2023learning,djeacoumar2025fractals},
the latter would mean that amortized initialization enters a lower-reconstruction-error basin more
easily, an advantage beyond speed. 

To test this, we extend refinement to $1000$
steps and compare the distribution of converged reconstruction quality. 
For random initialization we use both a single restart (random-r1) and the best of $4$
restarts (random-r4), and we align the budget by the number of optimization steps
(random-r4 costs four times the compute per step; this experiment measures
wall-clock time over all $256$ samples processed together, so the absolute seconds
per sample are not directly comparable to Section~\ref{subsec:pareto}).

\begin{figure}[t]
  \centering
  \includegraphics[width=\linewidth]{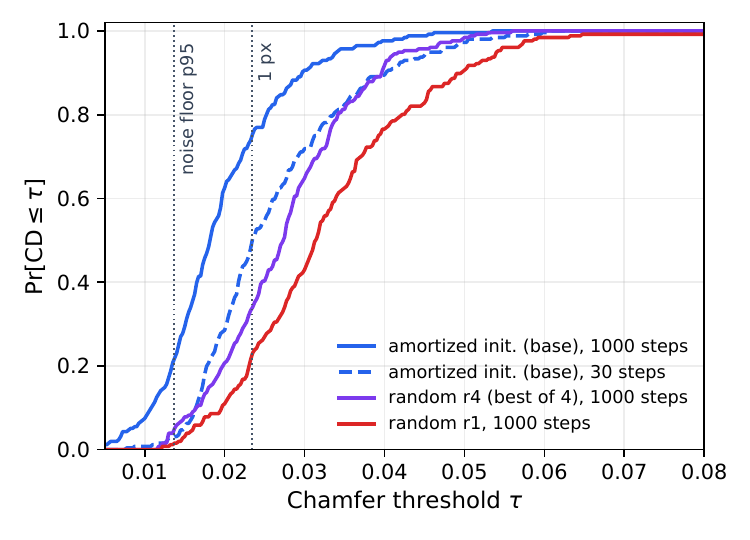}
  \caption{Success rate $\Pr[\mathrm{CD}\le\tau]$ as a function of the threshold
  $\tau$ (\texttt{test256}). Solid lines: $1000$ steps; dashed: \texttt{base} at
  $30$ steps. Dotted lines mark the $95$th percentile of the Chamfer noise floor
  (true parameters re-rendered) and one pixel width. At every $\tau$, amortized
  initialization is at least as successful as either random baseline with the same
  number of steps; this holds at each evaluated budget ($0$, $30$, $100$, $300$, and
  $1000$ steps), of which $1000$ is shown.}
  \label{fig:success_tau}
\end{figure}

In addition to the distribution, we use a success rate $\Pr[\mathrm{CD}\le\tau]$,
the fraction of samples whose converged Chamfer falls at or below a threshold
$\tau$. 
We fix the threshold independently of any method at the width of one pixel of the
$128\times128$ grid, $\tau=3/128\approx0.0234$; with the Chamfer distance of
Section~\ref{subsec:setup}, $\mathrm{CD}\le\tau$ means a root-mean-square
nearest-neighbor distance of at most $\sqrt2$ pixels. For reference, re-rendering the true
parameters under the same protocol gives a Chamfer noise floor of $0.0068$ on average
($95$th percentile $0.0137$), so $\tau$ is about $3.4$ times the mean floor. Because any single threshold is a
choice, we also report the success rate over the whole range of $\tau$
(Figure~\ref{fig:success_tau}).

\begin{table*}[t]
  \centering
  \small
  \caption{Convergence distribution under long optimization (\texttt{test256},
  matched renderer). Converged values at $1000$ steps for \texttt{base} (amortized
  initialization) and random initialization (r1 / best of r4). The CD columns are
  the mean, median, and $p_{95}$ of Chamfer; density SSE is the mean (all better
  when smaller). The success rate is $\Pr[\mathrm{CD}\le\tau]$ with $\tau$ equal to
  one pixel width ($3/128\approx0.0234$). +SA: gradient descent interleaved with
  simulated-annealing candidate search in the style of Djeacoumar et
  al.~\cite{djeacoumar2025fractals}, re-implemented in our objective. All $256$
  samples are refined in one batch; since random rendering samples and gradient
  clipping are per batch, the \texttt{base} rows differ slightly from
  Table~\ref{tab:hero_pareto} ($0.0259$ at $30$ steps, batches of $32$) and
  Section~\ref{subsec:training_signal} ($0.0190$ at $1000$ steps).}
  \label{tab:convergence}
  \begin{tabular}{llrrrrrr}
    \toprule
    Method & step & density SSE & CD mean & CD median & CD $p_{95}$ & success$_{1\,\mathrm{px}}$ & $(W,b)$ err \\
    \midrule
    \texttt{base}      & 30   & $2.27\times10^{-4}$ & 0.0262 & 0.0234 & 0.0457 & 0.504 & 0.395 \\
    \textbf{\texttt{base}} & \textbf{1000} & $\mathbf{1.55\times10^{-4}}$ & \textbf{0.0195} & \textbf{0.0181} & \textbf{0.0340} & \textbf{0.754} & 0.453 \\
    \midrule
    random-r1          & 1000 & $2.82\times10^{-4}$ & 0.0331 & 0.0314 & 0.0540 & 0.230 & 0.971 \\
    random-r4          & 1000 & $2.09\times10^{-4}$ & 0.0276 & 0.0273 & 0.0417 & 0.340 & 0.928 \\
    \midrule
    \texttt{base} + SA  & 1000 & $1.48\times10^{-4}$ & 0.0192 & 0.0184 & 0.0342 & 0.773 & 0.451 \\
    random-r4 + SA     & 1000 & $1.87\times10^{-4}$ & 0.0256 & 0.0249 & 0.0392 & 0.418 & 0.882 \\
    \bottomrule
  \end{tabular}
\end{table*}

\begin{figure*}[t]
  \centering
  \includegraphics[width=0.48\linewidth]{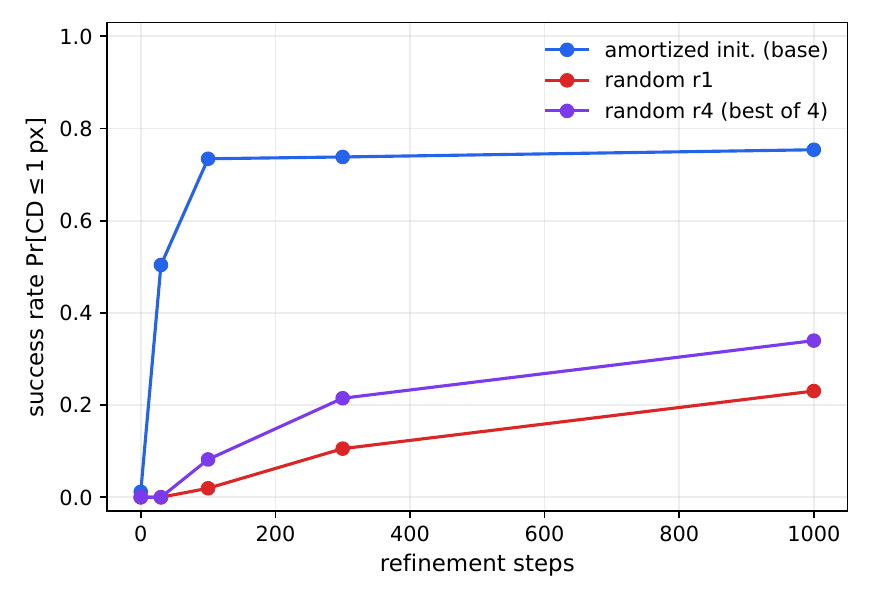}\hfill
  \includegraphics[width=0.48\linewidth]{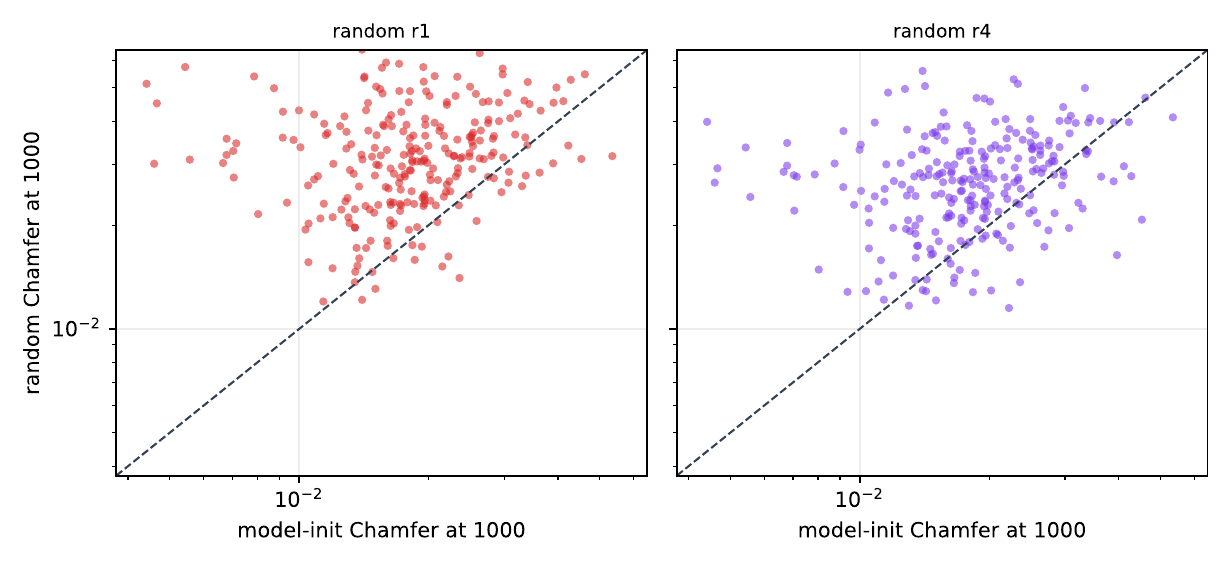}
  \caption{Convergence under long optimization (\texttt{test256}). Left: success
  rate $\Pr[\mathrm{CD}\le1\,\mathrm{px}]$ versus number of steps.
  \texttt{base} (amortized initialization) reaches $0.73$ by $100$ steps and
  saturates at about $0.75$, whereas random initialization remains at $0.23$ (r1) / $0.34$
  (r4) even at $1000$ steps. Right: pairwise scatter of the $1000$-step converged
  Chamfer for the same samples (horizontal \texttt{base}, vertical random). Almost
  all points lie above the diagonal, so amortized initialization converges to a
  better solution sample by sample.}
  \label{fig:convergence}
\end{figure*}

As Table~\ref{tab:convergence} and Figure~\ref{fig:convergence} show, even extended
to $1000$ steps the converged quality of random initialization does not reach that
of amortized initialization (mean Chamfer: random-r4 $0.0276$, random-r1 $0.0331$
vs \texttt{base} $0.0195$). The gap is larger in the tail of the distribution
($p_{95}$: random-r1 $0.0540$, random-r4 $0.0417$ vs \texttt{base} $0.0340$),
showing that some samples stall at poor solutions. HD95 and coverage@2px
confirm the same ordering (HD95: \texttt{base} $1000$ at $0.070$ vs random-r4
$0.105$; coverage: $0.920$ vs $0.861$). The gap is smallest on density SSE
(\texttt{base} $1000$ $1.55\times10^{-4}$ vs random-r4 $2.09\times10^{-4}$), where
random-r4 even falls below \texttt{base}+30 ($2.27\times10^{-4}$) on average;
because the refinement objective is dominated by the density term, random
initialization can lower the directly-optimized density while converging to a basin
whose shape (Chamfer/HD95/coverage) does not match (the decoupling of
reconstruction objective, shape, and parameters; Section~\ref{sec:identifiability}).
By the success rate at one pixel width the gap is substantial: \texttt{base} reaches $0.504$ at $30$ steps and $0.754$ at $1000$ steps,
whereas random initialization stays at $0.230$ (r1) and $0.340$ (r4) even at $1000$
steps (Figure~\ref{fig:convergence}, left). Taking the best of $4$ restarts and
spending four times the compute per step does not close this gap. The ordering does
not depend on the threshold (Figure~\ref{fig:success_tau}): at each evaluated budget
($0$, $30$, $100$, $300$, and $1000$ steps) and for every $\tau\in[0.005,0.08]$, the
success rate of amortized initialization is at least that of either random baseline
with the same number of steps, i.e., its Chamfer distribution stochastically
dominates theirs.
With a per-sample criterion that uses no global threshold,
$\mathrm{CD}_i\le3\,\mathrm{floor}_i$ with $\mathrm{floor}_i$ the noise floor of
sample $i$, the $1000$-step rates are $0.590$, $0.297$, and $0.195$ for \texttt{base},
random-r4, and random-r1. Moreover,
\texttt{base}+30 (mean $0.0262$) matches random-r4 at $1000$ steps ($0.0276$),
reaching comparable quality with $33\times$ less optimization. The pairwise scatter
(Figure~\ref{fig:convergence}, right) confirms this per sample: amortized
initialization converges to a better solution on almost every sample.

Djeacoumar et al.~\cite{djeacoumar2025fractals} escape poor local minima by
interleaving gradient descent with simulated-annealing candidate search. We
re-implemented this hybrid in our objective, following their schedule (every
$K/60$ steps during the first half of a $K$-step budget, $10$ Gaussian candidates
with a linearly decreasing temperature and Metropolis acceptance) but perturbing our
raw $(W,b)$ and measuring the energy change relative to the current objective. It
helps random initialization (random-r4 at $100/300/1000$ steps:
$0.0416/0.0311/0.0276\to0.0289/0.0266/0.0256$; success at $1000$ steps
$0.340\to0.418$) but leaves amortized initialization nearly unchanged
($0.0195\to0.0192$). The gap between the two remains
(Table~\ref{tab:convergence}): search narrows, but does not remove, the
disadvantage of a poor start. Their official code is compared on \texttt{test256}
systems in Section~\ref{subsec:external}.

Notably, \texttt{base}'s refinement lowers Chamfer substantially ($0.055\to0.019$)
while the $(W,b)$ error instead increases ($0.405\to0.453$)
(Table~\ref{tab:convergence}): convergence to a low-reconstruction-error solution is not convergence to the true parameters (Sections~\ref{sec:identifiability}
and~\ref{sec:discussion}). Random initialization has both worse reconstruction and a
high $(W,b)$ error (about $0.93$--$0.97$), falling into a different basin that does
not even reach a good reconstruction.

\subsection{Choice of training signal}
\label{subsec:training_signal}
Section~\ref{subsec:training} argued that matching the generating parameters is
preferable to training on reconstruction because refinement from a parameter-matched
prediction reaches better minima. We test this, together with the set-matching formulation itself,
by retraining the estimator with the encoder, data, and $50{,}000$-step budget of
\texttt{base} (Table~\ref{tab:training_signal}). The variants differ from
\texttt{base} in the training signal and, where stated, in the head, its
initialization, the training budget, or the checkpoint selection.
\begin{itemize}
  \item \emph{Hungarian, no fixed point}: \texttt{base} with $\lambda_{\mathrm{fp}}=0$.
  \item \emph{Ordered regression}: the canonical-ordering principle
  of~\cite{graham2021nnifs} on our backbone. The ground-truth maps are sorted by
  their fixed points ($x$, then $y$), and output slot $i$ is regressed on the $i$-th
  map without the fixed-point term, so that this variant differs from the previous
  one only in the matching.
  \item \emph{Multi-hypothesis}: the head outputs $K{=}4$ candidate sets and is
  trained with relaxed winner-takes-all~\cite{rupprecht2017multiple}; the candidate
  with the lowest matching loss receives weight $0.95$ and the others share $0.05$.
  At inference, the candidate with the lowest rendered density error against the
  observation is kept, without ground truth.
  \item \emph{\texttt{+aux}}: a $3{,}000$-step continuation of \texttt{base} that
  adds a reconstruction auxiliary (density error with weight $4$ and the squared
  Chamfer term of Section~\ref{subsec:inference} with weight $1$) to the matching loss.
  \item \emph{Reconstruction only}: no ground truth is used; the estimator minimizes
  the loss of the refinement objective of Section~\ref{subsec:inference} (the same
  terms and weights, with fresh rendering samples at every step) through the
  differentiable renderer. The $n$ output maps are initialized at distinct fixed
  points to break their symmetry, and the checkpoint is selected by validation
  reconstruction.
  \item \emph{Collage only}: neither ground truth nor rendering is used; the estimator
  minimizes the collage error of the Collage Theorem~\cite{barnsley1988fractals} for
  the observed measure, $\|y-\sum_i p_i\,(f_i)_{\#}\,y\|^2$ on the pixel grid with
  $p_i\propto|\det W_i|$, plus $0.1$ times the squared Chamfer term of
  Section~\ref{subsec:inference} between points drawn from $y$ and their images under
  the maps. The output is constrained to
  $s_{\max}\le0.70$, the upper end of the training prior, by the rescaling of
  Section~\ref{subsec:arch}; a version with only the penalties of
  reconstruction-only training is in Appendix~\ref{sec:appendix_training_signal}.
  Unlike the reconstruction-only variant, it is evaluated at the end of training:
  selecting the checkpoint by validation reconstruction would pick the untrained
  initialization, whose density error happens to be the lowest.
\end{itemize}

\begin{table*}[t]
  \centering
  \small
  \caption{Choice of training signal (\texttt{test256}, matched renderer). All
  variants share the encoder and data of \texttt{base} and, except for \texttt{+aux},
  its training budget; their other differences are listed in
  Section~\ref{subsec:training_signal}.
  Chamfer (CD) after $0$, $30$, and $100$ refinement steps, HD95 and coverage@2px
  after $30$ steps, and the number of samples (of $256$) on which \texttt{base} has
  the lower Chamfer at $30$ steps. All budgets in
  Appendix~\ref{sec:appendix_training_signal}.}
  \label{tab:training_signal}
  \begin{tabular}{lcrrrrrr}
    \toprule
    Training signal & uses GT $\theta$ & CD$_{0}$ & CD$_{30}$ & CD$_{100}$ & HD95$_{30}$ & cov$_{30}$ & \texttt{base} better \\
    \midrule
    \texttt{base}: Hungarian + fixed point & yes & 0.0546 & 0.0259 & 0.0201 & 0.0913 & 0.862 & -- \\
    Hungarian, no fixed point & yes & 0.0535 & 0.0264 & 0.0202 & 0.0940 & 0.859 & 135 \\
    ordered (fixed-point order) & yes & 0.0852 & 0.0384 & 0.0250 & 0.1443 & 0.776 & 202 \\
    multi-hypothesis ($K{=}4$) & yes & 0.0459 & \textbf{0.0229} & \textbf{0.0192} & \textbf{0.0813} & \textbf{0.888} & 80 \\
    \texttt{+aux}: \texttt{base} + reconstruction & yes & 0.0527 & 0.0264 & 0.0207 & 0.0942 & 0.859 & 135 \\
    reconstruction only & no & \textbf{0.0340} & 0.0307 & 0.0299 & 0.1219 & 0.836 & 175 \\
    collage only, $s_{\max}\le0.70$ & no & 0.0796 & 0.0390 & 0.0359 & 0.1543 & 0.788 & 219 \\
    \bottomrule
  \end{tabular}
\end{table*}

Reconstruction-only training gives by far the best one-shot output (Chamfer
$0.0340$ vs $0.0546$, better than \texttt{base} on $223$ of $256$ samples), about
as good as $1000$ steps of random-start refinement (random-r1, $0.0331$;
Table~\ref{tab:convergence}). Refinement, however, barely improves it
($0.0340\to0.0307\to0.0299\to0.0295$ at $30$, $100$, and $1000$ steps, a $13\%$
reduction), so that after $1000$ steps it is worse even than random-r4 ($0.0276$).
Refinement from \texttt{base}, in contrast, keeps descending
($0.0546\to0.0259\to0.0201\to0.0190$, a $65\%$ reduction); \texttt{base} is ahead
from $20$ steps on and is better on $175$ of $256$ samples at $30$ steps. The reconstruction-only predictions are also far from the generating
parameters: their $(W,b)$ error is $1.07$, about the mean distance between two
unrelated test parameter sets ($1.06$; Table~\ref{tab:ident_dist}), against $0.40$
for \texttt{base}, although they remain contractive and non-degenerate (largest
singular value at most $0.83$, no coincident maps). This is the behavior anticipated
in Section~\ref{subsec:training}. Direct measurements (medians over \texttt{test256})
show that the difference lies not in whether refinement moves the predictions but in
where it takes them. Refinement moves the reconstruction-only predictions about as far
in parameter space as those of \texttt{base}, but to poorer solutions: the objective
falls only from $5.5\times10^{-4}$ to $4.0\times10^{-4}$ after $1000$ steps, against
$1.3\times10^{-3}$ to $2.2\times10^{-4}$ from \texttt{base}. Perturbing the
predictions with Gaussian noise (standard deviation $0.05$ or $0.1$) does not change
this: $100$ refinement steps return to Chamfer $0.0303$ and $0.0315$ from
reconstruction-only ($0.0299$ without noise) and to $0.0210$ and $0.0239$ from
\texttt{base} ($0.0201$), even where \texttt{base} starts from the worse
reconstruction. Reconstruction-only training thus places its predictions where
gradient descent converges to poorer minima, and parameter matching places them where
it converges to better ones.

Collage training, the classical rendering-free route to an IFS, is the weakest
variant: Chamfer $0.0796$ one-shot and $0.0390$ after $30$ steps, with \texttt{base}
better on $219$ of $256$ samples ($0.1011$ and $0.0411$ without the output
constraint; Appendix~\ref{sec:appendix_training_signal}). The estimator does minimize
what it is trained on: on \texttt{test256} the collage error of its predictions
($6.1\times10^{-4}$ on average) is close to that of the generating parameters
($3.4\times10^{-4}$) and below it on $20$ of $256$ images, whereas that of
\texttt{base} is five times the ground-truth value ($1.7\times10^{-3}$). The predicted maps overlap
heavily (median $\sum_i|\det W_i|$ of $1.95$, against $0.76$ for the generating maps),
and their attractors are far from the targets. The Collage Theorem bounds the attractor
error by the collage error divided by $1-s$ only in the Hausdorff metric (for measures,
the Hutchinson metric); averaged errors on a pixel grid and on samples, which are what
can be evaluated without rendering, carry no such bound, and here a small collage error
does not mean a good reconstruction.

The \texttt{+aux} continuation shows the same trade-off in a milder form. It improves
the one-shot output ($0.0546\to0.0527$; a continuation without the auxiliary does
not, $0.0552$) and moves the prediction slightly away from the generating parameters
($(W,b)$ error $0.405\to0.413$, GT-$\theta$ validation loss $0.169\to0.173$), and
after $30$ and $100$ steps it is tied with \texttt{base} ($135$ and $133$ of $256$
samples favor \texttt{base}, neither significant under a sign test). Without refinement,
reconstruction-only training is far better ($0.0340$), so \texttt{+aux} offers no
advantage at either budget, and we keep \texttt{base} as the standard model.
The fixed-point term is not essential: without it the one-shot output is slightly
better ($0.0535$) and the refined output slightly worse ($0.0264$), and the
per-sample comparison is close to even ($118$ and $135$ of $256$ samples favor
\texttt{base} at $0$ and $30$ steps).

The set-matching loss, in contrast, matters. Ordered regression, which differs from
Hungarian matching without the fixed-point term only in how predicted and true maps
are paired, is clearly worse: Chamfer $0.0852$ one-shot, $0.0384$ after $30$
steps, $0.0250$ after $100$, and $0.0217$ after $1000$ steps against $0.0535$,
$0.0264$, $0.0202$, and $0.0191$, with the Hungarian variant better on $218$, $203$, $181$,
and $160$ of $256$ samples, and a larger $(W,b)$ error ($0.485$ against $0.407$). A fixed canonical order assigns
nearly equivalent maps to different slots whenever two fixed points swap their order,
so the regression targets jump discontinuously and the predictions are pulled further
toward an average (mean largest singular value $0.33$, against $0.37$ with Hungarian
matching and $0.53$ for the true maps).

Multiple hypotheses address the ambiguity directly. The four candidate sets of one
image differ substantially (mean $(W,b)$ distance between candidates $0.39$, about the
error of \texttt{base}), and each candidate alone is about as good as \texttt{base}
(Chamfer $0.055$--$0.060$ per slot). Choosing among them by the rendered density error
against the observation recovers most of what an oracle choice would gain (in a
separate analysis with its own point samples, Chamfer $0.0458$ against $0.0432$ for the best of the four by ground truth and $0.0573$ for a
single slot on average) and improves on \texttt{base} one-shot ($0.0459$), after $30$
steps ($0.0229$; \texttt{base} better on $80$ of $256$ samples), and, by a narrowing
margin, after $100$ steps ($0.0192$ vs $0.0201$; $111$ of $256$); after $1000$ steps
the two are level ($0.0192$ vs $0.0190$; $124$ of $256$). The
candidate nearest the generating parameters is often not the one that reconstructs
best (they coincide for $44\%$ of the images), another manifestation of the non-identifiability
of Section~\ref{sec:identifiability}. The selection renders each candidate once, which
adds $2.4$\,ms per image in batches of $32$ on an idle RTX~3090 ($85$\,ms for a single
image, where rendering is not shared across a batch), against $0.45$\,ms for the forward
pass. We keep \texttt{base} as the standard model, since
all other experiments use it, and regard the multi-hypothesis head as a direct
extension that exploits the known renderer to resolve ambiguity at inference.

\subsection{Generalization (1): distribution shift and observation mismatch}
\label{subsec:ood}
We test generalization to fractals of the same family that lie outside the
training distribution (out of distribution, OOD).
Keeping the number of maps $n{=}4$ and the matched renderer as at training, we
shift the evaluation data outside the training prior in four ways: S1 (large
singular values) $s_1,s_2\sim\mathcal U(0.70,0.85)$ (above the training upper bound
$0.70$); T1 (wide translation) fixed points $\sim\mathcal U(-1.2,1.2)^2$ (beyond the
training $\pm0.75$); R (reflections), the training generator with the sign of each
singular value flipped at random, so that $55\%$ of the maps reverse orientation and
every system contains at least one; and P (independent probabilities), the
\texttt{test256} systems rendered with selection probabilities
$p=(1-\alpha)\,p_{\det}+\alpha\,q$, where $q$ is drawn from
$\mathrm{Dirichlet}(1,\dots,1)$, instead of $p_{\det}\propto|\det W|$. For each setting, with $256$ samples, we evaluate the
standard model at $0/10/20/30$ steps and equal-budget random-r4.

\begin{table*}[t]
  \centering
  \small
  \caption{Within-family OOD ($256$ samples each, Chamfer). Amortized $0/30$ steps
  and random-r4 (equal-time $30$ / double-time $60$). R: reflections; P:
  selection probabilities independent of $|\det W|$ ($\alpha{=}1$; other $\alpha$ in
  Appendix~\ref{sec:appendix_robustness}).}
  \label{tab:ood}
  \begin{tabular}{lrrrr}
    \toprule
    Shift & ours-0 & ours-30 & random-r4-30 & random-r4-60 ($2\times$) \\
    \midrule
    S1 (large singular values) & 0.1271 & 0.0466 & 0.0476 & \textbf{0.0349} \\
    T1 (wide translation)      & 0.0996 & \textbf{0.0481} & 0.0950 & 0.0685 \\
    R (reflections)            & 0.0606 & \textbf{0.0318} & 0.0758 & 0.0533 \\
    P (independent probabilities) & 0.0632 & \textbf{0.0354} & 0.0789 & 0.0577 \\
    \bottomrule
  \end{tabular}
\end{table*}

In all settings the $0$-step result is worse than in distribution ($0.0546$) but
recovers clearly with refinement. The behavior, however, differs by the type of
shift (Table~\ref{tab:ood}; quality--speed curves in
Appendix~\ref{sec:appendix_robustness}). For T1 (location shift),
amortized $30$-step not only exceeds random-r4 at equal time but also exceeds the
double-time random-r4-60 (per sample Chamfer $208/256$, HD95 $215/256$).
For S1, where larger singular values produce weaker contractions and more space-filling attractors than those seen in training, amortized refinement is roughly tied with random-r4 at equal time but is overtaken by random-r4-60.
The advantage thus depends on how much of the learned structural prior the shift
preserves: it persists under structure-preserving shifts such as T1, where only the
location changes, and fades when the geometry of the attractors itself is novel, as
in S1, where per-instance optimization becomes favorable given enough budget.

Reflections and independent probabilities, which the training prior excludes, behave
like T1. Under R, amortized $30$-step ($0.0318$) beats random-r4 at equal time on
$251$ of $256$ samples and also beats the double-time random-r4-60 ($0.0533$). The
estimator almost never predicts a reflecting map ($0.1\%$ of maps at $0$ steps), so
it reproduces reflected attractors with orientation-preserving maps, another
manifestation of non-identifiability. Including reflections in the training prior
does not help: the estimator then predicts reflecting maps at about the right rate
but reconstructs R worse, because the orientation of each map adds an ambiguity that
the regression averages over (Appendix~\ref{sec:appendix_robustness}). Under P the
$30$-step Chamfer rises gradually with $\alpha$, from $0.0258$ ($\alpha{=}0$) to
$0.0354$ ($\alpha{=}1$), still ahead of random-r4-60 at every $\alpha$.

The estimator is also trained on density maps from a single renderer
($16$ trajectories of $1024$ steps, hard histogram, smoothing $\sigma{=}2$, $128^2$
grid). To test sensitivity to how the observation is rendered, we keep the
\texttt{test256} systems and change one setting of the observation renderer at a time
(Table~\ref{tab:renderer}; full sweep in Appendix~\ref{sec:appendix_robustness}).
Refinement keeps the training-time renderer, i.e., the observation's renderer is
treated as unknown; refining with the observation's settings instead changes the
results little. Because the metrics compare point clouds of the predicted and true
attractors, they do not depend on the observation renderer.

\begin{table}[t]
  \centering
  \small
  \caption{Observation-renderer mismatch (\texttt{test256}, Chamfer). One setting of
  the observation renderer is changed at a time; refinement uses the training
  renderer. The last column counts the samples (of $256$) on which ours-30 beats
  random-r4-30. Observations and random initializations are redrawn, so the first
  row differs slightly from Table~\ref{tab:hero_pareto}.}
  \label{tab:renderer}
  \setlength{\tabcolsep}{3pt}
  \begin{tabular}{lrrrr}
    \toprule
    Observation & ours-0 & ours-30 & r4-30 & wins \\
    \midrule
    training renderer & 0.0546 & 0.0258 & 0.0726 & 250 \\
    $2$ trajectories & 0.0575 & 0.0273 & 0.0728 & 254 \\
    resolution $64^2$ & 0.0600 & 0.0305 & 0.0734 & 248 \\
    resolution $256^2$ & 0.0544 & 0.0256 & 0.0726 & 250 \\
    bilinear splatting & 0.0546 & 0.0258 & 0.0726 & 249 \\
    no burn-in & 0.0546 & 0.0257 & 0.0730 & 250 \\
    $\sigma{=}4$ & 0.0728 & 0.0423 & 0.0773 & 239 \\
    $\sigma{=}1$ & 0.1456 & 0.0756 & 0.0721 & 135 \\
    $\sigma{=}0$ & 0.2032 & 0.1166 & 0.0716 & 61 \\
    $\sigma{=}1$, pre-blurred by $2$ & 0.0604 & 0.0306 & 0.0730 & 250 \\
    \bottomrule
  \end{tabular}
\end{table}

The estimator is insensitive to the sampling of the observation: with only $2$
trajectories, a $64^2$ or $256^2$ grid resampled to $128^2$, bilinear splatting, or no
burn-in, the $30$-step Chamfer stays within $0.0256$--$0.0305$ and beats random-r4 on
$248$--$254$ of $256$ samples. It is sensitive to the smoothing width $\sigma$. A
larger $\sigma$ degrades the result moderately ($0.0423$ at $\sigma{=}4$, still $239$
wins), whereas a smaller width makes the one-shot prediction fail ($0.1456$ at
$\sigma{=}1$, $0.2032$ at $\sigma{=}0$), and refinement then ties with or falls behind
random-r4. Random-r4 itself is insensitive to $\sigma$ ($0.0716$--$0.0773$), so the
failure is a shift in the estimator's input distribution rather than a harder
problem. Since smoothing can always be added, a sharper observation can simply be
blurred before inference, the blurred image serving as both the input and the
refinement target: blurring a $\sigma{=}1$ observation by $\sigma{=}2$ restores
$0.0306$ and $250$ wins (Appendix~\ref{sec:appendix_robustness}).

\subsection{Generalization (2): an external benchmark and comparison with Djeacoumar et al.}
\label{subsec:external}
We next test the estimator on IFS that were not produced by our generator and compare
it with the state-of-the-art per-image optimizer. Djeacoumar et
al.~\cite{djeacoumar2025fractals} evaluate on $101$ random IFS generated with the
sampler of Anderson and Farrell~\cite{anderson2022improving} and publish the codes
optimized by their method (about an hour per image) and by three baselines: an
evolutionary algorithm, cuckoo search, and a CNN that regresses the parameters. These
systems lie outside our training prior on several axes at once: they have $3$--$10$
maps (median $5$), $92\%$ of them have a map with $s_{\max}>0.70$ (median of the
per-system maximum $0.88$), $87\%$ contain a reflection, and their renderer draws
maps with probabilities $\propto|\det W|+0.01$. Because their renderer normalizes each
attractor to its bounding box, we normalize the ground truth and each published code
to the same box in our domain, give our estimator the observation of the normalized
ground truth, and score all methods with our point-cloud metrics (details in
Appendix~\ref{sec:appendix_external}). We apply
our $n{=}10$ model (Section~\ref{subsec:mnist_tu}) without retraining, so every method uses $10$ maps.

\begin{table*}[t]
  \centering
  \small
  \caption{External benchmark of Djeacoumar et al.~\cite{djeacoumar2025fractals}
  ($101$ IFS). Chamfer mean and median, HD95, coverage@2px, success rate
  $\Pr[\mathrm{CD}\le1\,\mathrm{px}]$, the number of IFS on which the method has a
  lower Chamfer than Djeacoumar et al., the percentage of IFS with a non-contractive
  map ($s_{\max}\ge1$), and time per image (for ours and random-r4: the wall-clock time
  of batched runs divided by the number of images / a single image processed alone). Ours and random-r4 use our $n{=}10$ model
  and objective; the other rows are the codes published with the benchmark.
  $^\dagger$Reported by Djeacoumar et al.\ (NVIDIA A40); our times are on an idle
  RTX~3090, on which their official code took $50$ min per system
  (Appendix~\ref{sec:appendix_external}).}
  \label{tab:external}
  \setlength{\tabcolsep}{4pt}
  \begin{tabular}{lrrrrrrrr}
    \toprule
    Method & CD mean & CD median & HD95 & cov@2px & success$_{1\,\mathrm{px}}$ & better & non-contr.\ (\%) & time \\
    \midrule
    ours, $0$ steps    & 0.0590 & 0.0593 & 0.2438 & 0.696 & 0.00 & 2  & 0 & $0.001$ / $0.02$\,s \\
    ours, $30$ steps   & 0.0247 & 0.0248 & 0.0847 & 0.888 & 0.43 & 17 & 0 & $1.3$ / $18$\,s \\
    ours, $100$ steps  & 0.0207 & 0.0205 & 0.0690 & 0.918 & 0.83 & 27 & 0 & $4.2$ / $59$\,s \\
    ours, $1000$ steps & \textbf{0.0201} & 0.0197 & \textbf{0.0674} & 0.926 & \textbf{0.86} & 39 & 0 & $41$ / $590$\,s \\
    random-r4, $100$ steps  & 0.0264 & 0.0257 & 0.0968 & 0.889 & 0.28 & 15 & 0 & $4.2$ / $59$\,s \\
    random-r4, $1000$ steps & 0.0210 & 0.0208 & 0.0714 & 0.921 & 0.78 & 31 & 0 & $41$ / $590$\,s \\
    \midrule
    Djeacoumar et al.  & 0.0210 & \textbf{0.0178} & 0.0704 & \textbf{0.943} & 0.78 & -- & 0 & $\approx1$\,h$^\dagger$ \\
    evolutionary       & 0.0668 & 0.0363 & 0.2646 & 0.778 & 0.06 & 0 & 34 & -- \\
    cuckoo search      & 0.0916 & 0.0802 & 0.3639 & 0.626 & 0.00 & 1 & 71 & -- \\
    neural regression  & 0.1330 & 0.1313 & 0.4792 & 0.443 & 0.00 & 0 & 3 & -- \\
    \midrule
    ground truth (re-rendered) & 0.0108 & 0.0108 & 0.0293 & 0.992 & 1.00 & -- & 0 & -- \\
    \bottomrule
  \end{tabular}
\end{table*}

The one-shot prediction is far from the targets (Chamfer $0.0590$), as expected
for systems this far outside the prior, but refinement recovers quickly and the
amortized start keeps its advantage over random starts at every budget ($0.0207$ vs
$0.0264$ at $100$ steps; Table~\ref{tab:external}). Against Djeacoumar et al.\ the
comparison is two-sided. Their search, which takes about an hour per image, is more
accurate on most individual IFS ($62$ of $101$ against our $1000$-step result; median
$0.0178$ vs $0.0197$). It also fails on a minority, however ($15$ IFS above Chamfer
$0.03$, the worst at $0.075$), whereas our worst $1000$-step result is about $0.03$.
The means are therefore equal ($0.0210$ against $0.0207$ after $100$ steps, about
$1$\,min for a single image, and $0.0201$ after $1000$ steps, about $10$\,min), and our
success rate at one pixel width is higher ($0.83$ at $100$ steps vs $0.78$). Giving
their method and ours the scale and translation that best align each attractor with
the truth does not change this (Appendix~\ref{sec:appendix_external}). The three
baselines published with the benchmark are far behind, and two of them often produce
non-contractive maps. We also evaluated six hand-specified IFS, including the
Sierpi\'nski triangle and the Barnsley fern. They are not part of the benchmark, so no
results of the other methods exist for them; against random starts, our $n{=}4$ model
refined for $100$ steps beats random-r4 at $1000$ steps on all six (Chamfer $0.0242$
vs $0.0308$; Appendix~\ref{sec:appendix_external}).

Conversely, we ran their official code with its default configuration ($10$ maps,
$15{,}000$ iterations) on the first $16$ \texttt{test256} systems, each drawn by their
own target generator; it took $50$ min per system on our RTX~3090. These systems lie
inside our training prior, so this is the counterpart of the benchmark above, and here
our method is ahead. Their mean Chamfer is $0.0537$ (median $0.0359$), against $0.0283$
and $0.0211$ after $30$ and $1000$ steps from our estimate, which is better on $9$ and
$14$ of the $16$ systems. Checks of the alignment, of the number of maps, and of our
use of their code account for at most a small part of this gap
(Appendix~\ref{sec:appendix_external}). Rather, strong contraction appears to be hard
for their method: their published results fail most often on the benchmark systems
with $s_{\max}<0.75$, and every \texttt{test256} system contracts this strongly. With
$16$ systems, this comparison is indicative rather than conclusive.

\subsection{Generalization (3): real images and comparison with Tu et al.}
\label{subsec:mnist_tu}
Finally, we compare transfer to non-fractal real images directly against the
per-image optimization of Tu et al.\ (2023)~\cite{tu2023learning}, which has a
public implementation. We use two datasets, MNIST and Fashion-MNIST: the former is
close to occupancy (thin strokes), while the latter contains filled regions and
requires reproducing density. Here we use a scaled version with more maps: an
estimator trained with the same framework and data generation at $n{=}10$
($100{,}000$ steps, with the cosine decay starting at $60{,}000$ steps; otherwise
identical to Section~\ref{sec:method}) applied to both datasets. 
This also matches the number of maps to Tu's public setting ($n{=}10$).
For fairness, we render the IFS output by both methods under a common condition and
measure two families of metrics.
We refer to these as the density and occupancy conditions below.

\begin{description}
  \item[(i) Our density condition] A density map proportional to
    visit frequency on the matched renderer, evaluated with the four
    reconstruction metrics of Section~\ref{subsec:setup} (density SSE, Chamfer,
    HD95, and coverage@2px).
  \item[(ii) Tu's occupancy condition] Tu's renderer overlays an RBF kernel on each
    point and then clamps the pixel value to $[0,1]$. Because pixels where many
    points accumulate saturate at $1$, the resulting $32\times32$ image effectively
    represents occupancy (whether each region is covered by points) rather than
    visit frequency (density). On this occupancy image, the evaluation value is the
    minimum MSE over $100$ sampling sequences (occupancy MSE; Tu's evaluation
    procedure).
\end{description}
Tu's optimization was run with the public code for $n\in\{4,10\}$ ($1000$ iterations
per image). We additionally evaluate a hybrid denoted ours-init$+$occupancy-GD,
which starts from our estimator's one-shot output and optimizes Tu's occupancy
objective by gradient descent for $100$ steps.

\begin{table*}[t]
  \centering
  \small
  \caption{MNIST (balanced $50$ images, $n{=}10$ estimator). occupancy MSE is the
  minimum MSE under the occupancy condition; the rest are density metrics.
  density SSE, Chamfer, HD95, and occupancy MSE are better when smaller,
  coverage@2px when larger.}
  \label{tab:mnist_tu}
  \begin{tabular}{lrrrrr}
    \toprule
    Method & occupancy MSE & density SSE & Chamfer & HD95 & coverage@2px \\
    \midrule
    ours $0$-step                    & 0.0874 & $4.30\times10^{-4}$ & 0.0571 & 0.2245 & 0.6888 \\
    ours $30$-step                   & 0.0867 & $1.01\times10^{-4}$ & 0.0221 & 0.0753 & 0.9126 \\
    ours $100$-step                  & 0.0724 & $\mathbf{6.95\times10^{-5}}$ & \textbf{0.0193} & \textbf{0.0598} & \textbf{0.9357} \\
    ours-init $+$ occupancy-GD $100$ & 0.0190 & $7.67\times10^{-4}$ & 0.0485 & 0.1931 & 0.7114 \\
    Tu ($n{=}4$)                     & 0.0296 & $2.88\times10^{-3}$ & 0.1329 & 0.5305 & 0.6154 \\
    Tu ($n{=}10$)                    & 0.0214 & $4.91\times10^{-3}$ & 0.1017 & 0.3837 & 0.6364 \\
    \bottomrule
  \end{tabular}
\end{table*}

\begin{figure*}[t]
  \centering
  \begin{minipage}[t]{0.49\linewidth}
    \centering
    \includegraphics[width=\linewidth,trim=0 548 0 0,clip]{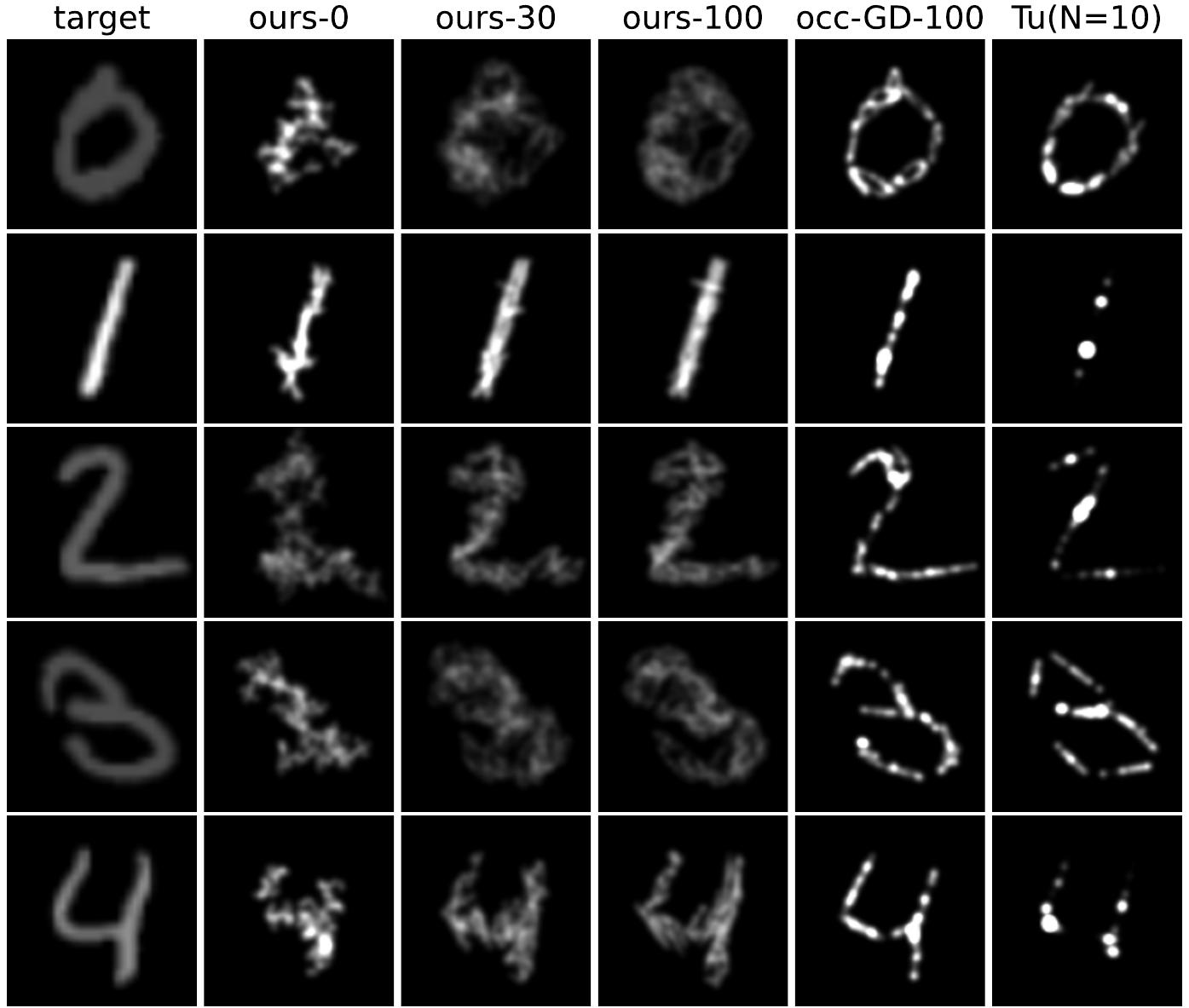}\\[2pt]
    {\small (a) MNIST}
  \end{minipage}\hfill
  \begin{minipage}[t]{0.49\linewidth}
    \centering
    \includegraphics[width=\linewidth,trim=0 548 0 0,clip]{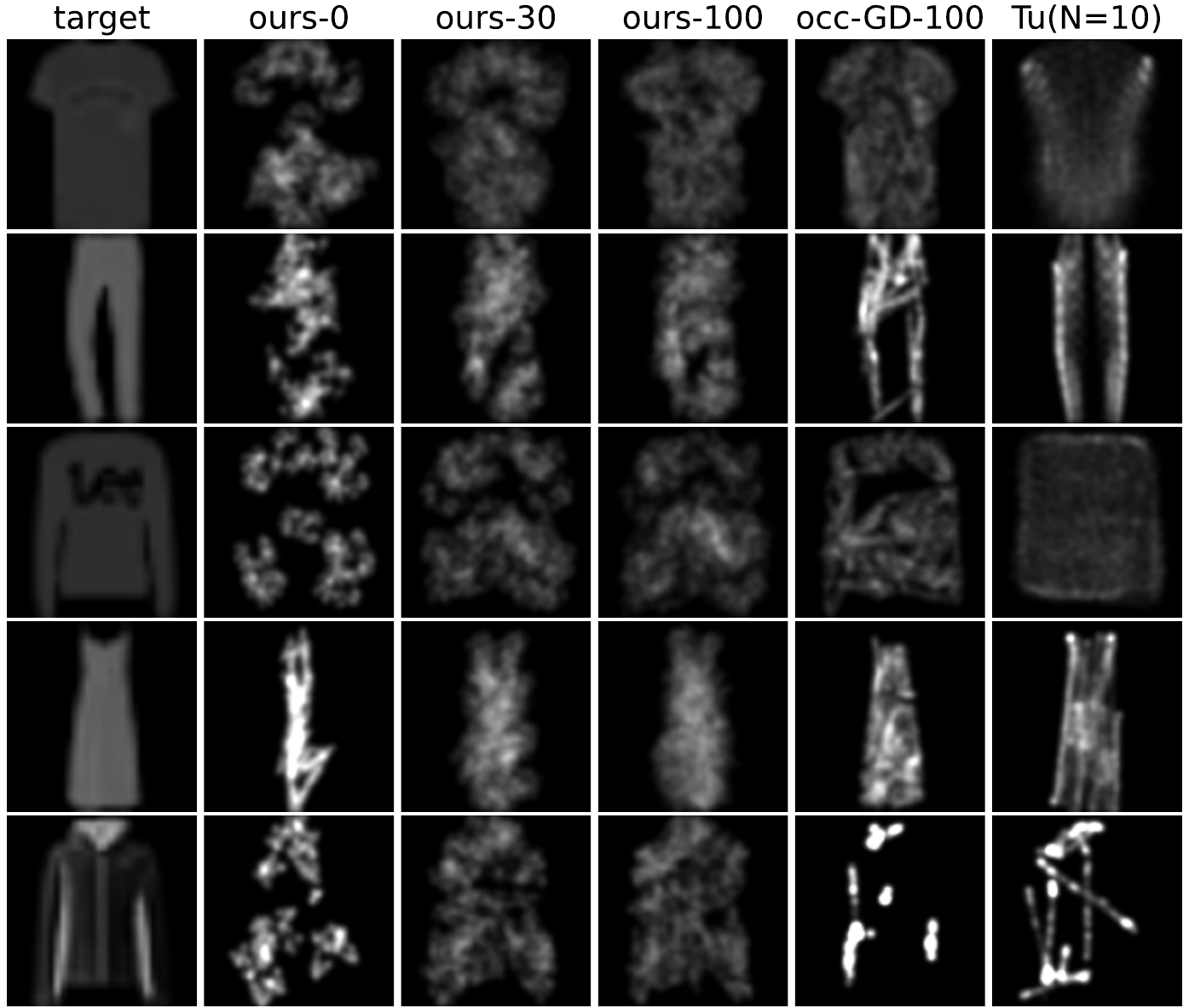}\\[2pt]
    {\small (b) Fashion-MNIST}
  \end{minipage}
  \caption{Reconstruction of real images (density maps, common grayscale, one example
  per row; the first five of the ten examples in Figures~\ref{fig:mnist_density}
  and~\ref{fig:fmnist_density}). Columns from left: target, ours $0$-step, ours
  $30$-step, ours $100$-step, ours-init$+$occupancy-GD $100$, Tu ($n{=}10$). Under
  density-aware refinement, mass spreads as an area and approaches the target's
  density as $0\to30\to100$ steps, whereas Tu and occupancy-GD concentrate mass on
  thin lines and isolated points (a), and occupancy-GD collapses the garments to
  outlines (b).}
  \label{fig:real_density}
\end{figure*}

The result depends on metrics (Table~\ref{tab:mnist_tu}, Figure~\ref{fig:real_density}a;
Appendix~\ref{sec:appendix_qualitative} gives ten examples per dataset, the same
outputs in Tu's occupancy view, and all $50$ images).
Under the density condition, our refined outputs are consistently better: ours at $30$ steps improves over Tu at $n{=}4,10$ on all $50$ images and all main metrics (e.g., Chamfer
$0.0221$ vs $0.1017$ for Tu-$n{=}10$), and improves further at $100$ steps. As
Figure~\ref{fig:real_density}a shows, Tu tends to concentrate mass on thin lines
and, as is notable for the digits $1,2,4$, sometimes on isolated points, whereas
our method tends to reconstruct mass as a filled area. 
Under the occupancy condition, our raw output is worse than Tu's, but ours-init$+$occupancy-GD reaches a lower mean occupancy MSE than Tu-$n{=}10$;
the per-sample win rate is only $20/50$, so this is not a uniform improvement.

On speed, Tu took about $100$--$160$ s per image. Processing one image at a time, our method
took about $0.3$ s at $0$ steps, $18$ s at $30$ steps, and $59$ s at the highest-quality
$100$ steps (about $350$--$550\times$, $6$--$9\times$, and $2$--$3\times$ faster,
respectively); in batches of ten images the times per image are $0.04$, $2.5$, and
$8.2$ s (about $2600\times$, $40$--$60\times$, and $12$--$20\times$). Occupancy-GD lowers occupancy MSE while
worsening density SSE (losing $0/50$ to ours-$30$ on the density metrics),
showing that the occupancy objective and the density objective are
incompatible. In short, increasing the number of maps to $n{=}10$ makes the method
competitive with per-image optimization on real images: it retains the advantage
in density reconstruction and speed, while saturated occupancy continues to favor
the per-image optimizer.

Fashion-MNIST provides the complementary case: garments are filled regions whose
faithful reconstruction requires reproducing the internal mass distribution
(density).
Applying the same $n{=}10$ estimator to Fashion-MNIST (balanced $50$ images), we
compared with Tu ($n{=}10$, $1000$ iterations per image) (Table~\ref{tab:fmnist},
Figure~\ref{fig:real_density}b).

\begin{table*}[t]
  \centering
  \small
  \caption{Fashion-MNIST (balanced $50$ images, $n{=}10$). occupancy MSE is the
  minimum MSE under the occupancy condition; the rest are density metrics.}
  \label{tab:fmnist}
  \begin{tabular}{lrrrrr}
    \toprule
    Method & occupancy MSE & density SSE & Chamfer & HD95 & coverage@2px \\
    \midrule
    ours $0$-step                    & 0.1087 & $1.38\times10^{-4}$ & 0.0503 & 0.2102 & 0.7724 \\
    ours $30$-step                   & 0.0997 & $3.98\times10^{-5}$ & 0.0251 & 0.0738 & 0.8846 \\
    ours $100$-step                  & 0.0966 & $\mathbf{3.09\times10^{-5}}$ & \textbf{0.0235} & \textbf{0.0652} & \textbf{0.8979} \\
    ours-init $+$ occupancy-GD $100$ & \textbf{0.0450} & $4.94\times10^{-4}$ & 0.0629 & 0.2554 & 0.7645 \\
    Tu ($n{=}10$, $1000$ iters)      & 0.0672 & $1.43\times10^{-4}$ & 0.0453 & 0.1754 & 0.8610 \\
    \bottomrule
  \end{tabular}
\end{table*}

On the density metrics, ours at $100$ steps is again best on average (density SSE
$3.09\times10^{-5}$ against $1.43\times10^{-4}$ for Tu-$n{=}10$), but the margin is
smaller than on MNIST: in filled regions, covering a region and spreading mass over
it nearly coincide, so even Tu, which optimizes occupancy, keeps the density
reasonably well. Ours-$100$ beats Tu-$n{=}10$ on $28$, $32$, and $30$ of the $50$
images in density SSE, Chamfer, and HD95, an average advantage rather than a uniform
one. Reproducing the internal mass still requires density optimization:
ours-init$+$occupancy-GD beats Tu-$n{=}10$ under the occupancy condition ($0.0450$ vs
$0.0672$, $48/50$) but degrades the density (density SSE $4.94\times10^{-4}$) and
collapses the garments to thin outlines (Figure~\ref{fig:real_density}b).

\section{Discussion}
\label{sec:discussion}

\subsection{Parameters as the training target, reconstruction as the criterion}
\label{subsec:disc_target}
Reconstruction quality and parameter error diverge throughout the experiments.
Refinement improves Chamfer while leaving the $(W,b)$ error nearly unchanged, or even increasing it under long optimization, and reconstruction-oriented training, as an auxiliary or on its own, improves the one-shot output while moving the prediction away from the generating parameters.
These results justify treating $(W,b)$ error as a diagnostic rather than as the success criterion.

The parameters nevertheless matter as a training target. The estimator does not recover them, but they mark a solution at the reconstruction noise floor: predictions trained to match them remain improvable by refinement, whereas refinement barely improves predictions trained on reconstruction alone (Section~\ref{subsec:training_signal}); random-start refinement likewise often stalls at poor solutions (Section~\ref{subsec:convergence}).
The choice of training signal therefore follows the refinement budget: without refinement, reconstruction-only training gives the best output (Chamfer $0.0340$ against $0.0546$, about as good as $1000$ steps of random-start refinement), whereas from $20$ steps on, parameter matching is better and keeps improving (Table~\ref{tab:appendix_training_signal}).
Where an observation leaves several parameter sets plausible, a single prediction must compromise among them; predicting several candidates and selecting one with the known renderer avoids this compromise and improves reconstruction (Section~\ref{subsec:training_signal}).

\subsection{Specifying the reconstruction measure}
\label{subsec:disc_measure}
Choosing reconstruction as the criterion leaves open which reconstruction measure to use, and three aspects of this choice affect the results.
First, the renderer that defines the objective must match the one that generated the target; otherwise a solution other than the true parameters can reach a lower loss (Section~\ref{sec:oracle}).
This requirement concerns the objective, not the input of the estimator, which is insensitive to how the observation is sampled and sensitive only to its smoothing width, so a sharper observation can be blurred to match (Sections~\ref{subsec:ood} and~\ref{subsec:limitations}).
The point-cloud metrics compare attractors rather than rendered images and so do not depend on the renderer, which allows us to score methods that render differently (Section~\ref{subsec:external}).
Second, valid objectives need not agree: occupancy and visit-frequency density converge to different solutions, and optimizing occupancy from our output improves the occupancy metric but worsens the density (Section~\ref{subsec:mnist_tu}).
Which objective is appropriate depends on the application, occupancy when coverage matters and density when the intensity within filled regions matters; the present work targets density.
Third, a rendering-free surrogate is no substitute.
Training on the collage error, the classical criterion, brings the collage error of the predictions close to that of the generating parameters yet is the weakest of the training signals compared (Section~\ref{subsec:training_signal}): the Collage Theorem bounds the attractor error in the Hausdorff metric, or for measures in the Hutchinson metric, but not in the pixel- or sample-averaged errors that can be computed without rendering.

The Hutchinson metric is the Wasserstein-1 distance, and a density map is a finite-sample approximation of the invariant measure.
Optimal-transport (OT) distances between measures could therefore serve both as a reconstruction loss in place of the pixel SSE and point-cloud Chamfer used here and as a collage loss that keeps the bound of the Collage Theorem without rendering; we leave this to future work.

\subsection{Conditions under which amortized inference is effective}
As the main comparison shows, the trained estimator acts not only as a fast
one-shot predictor but also as a useful initializer for per-instance
optimization, outperforming random-from-scratch optimization at equal and even
double budget. The range in which this advantage holds depends,
as the OOD experiments show, on how much the test distribution preserves the
structural prior learned during training.
In other words, amortized inference is most effective when the test distribution reuses structure present in the training distribution.
Even when several assumptions of the prior are violated at once, as on the external benchmark (Section~\ref{subsec:external}), the prediction remains a useful initialization: refinement from it matches the mean accuracy of a state-of-the-art per-image optimizer with fewer failures, although that optimizer reaches a closer fit on most individual systems.
At the same time, increasing the number of maps to $n{=}10$ makes the method
competitive with per-image optimization on real images
(Section~\ref{subsec:mnist_tu}), showing that the framework is not specific to
$n{=}4$.

\subsection{Limitations}
\label{subsec:limitations}
\begin{itemize}
  \item \textbf{Non-identifiability near the noise floor is undetermined.} We
        quantified non-identifiability down to an achievable tolerance (density
        $L_2\approx0.013$; Section~\ref{sec:identifiability}). Whether it
        persists as the tolerance shrinks to the reconstruction noise floor
        ($\sim2.7\times10^{-6}$ in density SSE; Section~\ref{sec:oracle}) is
        undetermined; settling it would require a second-solution analysis that
        drives a solution far from the truth down to the floor, which we leave
        for future work.
  \item \textbf{The selection probability is fixed to $p\propto|\det W|$.} The
        framework cannot represent an IFS whose probabilities are given
        independently of the determinant; for such targets the reconstruction
        degrades gradually with the deviation from $p_{\det}$ while remaining
        better than random-start optimization (Section~\ref{subsec:ood}).
        Estimating variable selection probabilities is left for future work.
  \item \textbf{The number of maps is fixed per model.} We used $n{=}4$ for the
        controlled core study and $n{=}10$ for the real-image and scaling
        experiments; handling a variable number of maps within a single model is
        outside the scope of this work. If the probability learning above were
        achieved, the model could learn an effective number of active maps adaptively.
  \item \textbf{The training prior excludes reflections.} Reflected attractors are
        still reconstructed, with orientation-preserving maps ($30$-step Chamfer
        $0.0318$ against $0.0258$ in distribution; Section~\ref{subsec:ood}), but
        adding reflections to the training prior does not help
        (Appendix~\ref{sec:appendix_robustness}); a multi-hypothesis head
        (Section~\ref{subsec:training_signal}) may handle the added ambiguity.
  \item \textbf{The observation must match the training smoothing width.} The
        estimator tolerates changes in the sampling and resolution of the
        observation but not in its smoothing width: a sharper observation
        ($\sigma<2$) makes the one-shot prediction fail, and a broader one
        degrades it (Section~\ref{subsec:ood}). A sharper observation must
        therefore be blurred before inference, and the blurred image then also
        serves as the refinement target.
\end{itemize}

\section{Conclusion}
\label{sec:conclusion}

We studied inverse IFS reconstruction as amortized set prediction from density maps. 
Because density maps do not uniquely determine IFS parameters, the method is evaluated and refined by reconstruction, yet trained by matching the generating parameters: reconstruction-only training gives a better one-shot output, but refinement from it converges to poorer minima, whereas refinement from parameter-matched predictions keeps improving.
The known renderer supplies unlimited labeled training pairs and an image-only refinement objective.

On in-distribution data, the amortized prediction followed by light refinement gives a better quality--speed trade-off than random-start per-image optimization, even when the latter receives twice the time. Long-horizon experiments show that the advantage also reflects more reliable convergence. The benefit persists under structure-preserving distribution shifts and under changes in how the observation is sampled, and weakens under stronger geometric shifts.
On an external benchmark that violates several assumptions of the training prior at once, refinement from the prediction matches the mean accuracy of an hour-long optimizer within a minute and with fewer failures, although that optimizer fits most individual systems more closely.
On real images, with the number of maps increased to ten, the method is stronger on average than the per-image optimizer of Tu et al.~\cite{tu2023learning} under density criteria, while that optimizer remains preferable under saturated occupancy.

Beyond the specific extensions discussed in
Section~\ref{subsec:limitations}, the same strategy may apply to inverse
problems with inexpensive simulators: train an amortized estimator on synthetic
pairs, use the simulator for reconstruction-based refinement, and evaluate by
the observable rather than by non-identifiable latent parameters.


\printcredits

\section*{Declaration of competing interest}
The author declares that there are no known competing financial interests or
personal relationships that could have appeared to influence the work reported in
this paper.

\section*{Funding}
This research did not receive any specific grant from funding agencies in the
public, commercial, or not-for-profit sectors.

\section*{Declaration of generative AI and AI-assisted technologies in the manuscript preparation process}
During the preparation of this work, the author used Claude (Anthropic) and ChatGPT (OpenAI) to
 improve the language and readability of the manuscript,
and assist in developing and debugging the source code used in the experiments.
All AI-assisted text was reviewed and edited by the author,
and all AI-assisted code was reviewed, modified, and tested by the author before use.
The author takes full responsibility for the content of the publication,
the correctness of the implementation, and the reported results.

\section*{Data availability}
The source code, trained models, and evaluation scripts required to reproduce all
results in this paper are publicly available at
\url{https://github.com/cncs-fit/amortized-ifs}.

\bibliographystyle{model1-num-names}

\bibliography{references}

\appendix
\setcounter{dbltopnumber}{3}
\renewcommand{\dbltopfraction}{0.95}
\numberwithin{figure}{section}
\numberwithin{table}{section}

\section{Contractivity of predicted and refined maps}
\label{sec:appendix_contractivity}
The output parameterization (Eq.~\eqref{eq:affine_output}) bounds the entries of
$W_i$ but not its largest singular value. Table~\ref{tab:contractivity} reports, for
each setting and number of refinement steps, the mean and maximum of
$s_{\max}(W_i)$ over all predicted maps and the percentage of maps with
$s_{\max}>0.9$ and with $s_{\max}\ge1$ (a contractivity violation). The one-shot
output never violates contractivity, nor does refinement up to $100$ steps.
Violations are rare and small: in the $1000$-step runs of the training-signal
comparison (Appendix~\ref{sec:appendix_training_signal}), one map of one test system
exceeds one for four of the five estimators (largest $1.031$), and further
optimization of our output under Tu et al.'s occupancy objective, which has no
feasibility terms, gives one violation on Fashion-MNIST. The estimator with
spectrally constrained output (Section~\ref{subsec:arch}), whose predicted linear
parts are rescaled to $s_{\max}\le0.95$, performs on par with \texttt{base} in Chamfer
distance from $0$ to $100$ refinement steps (Table~\ref{tab:appendix_training_signal}).

\begin{table*}[t]
  \centering\small
  \caption{Largest singular values of the maps ($n{=}4$ for \texttt{test256}, T1,
  S1, R, P, and the hand-specified IFS; $n{=}10$ for MNIST, Fashion-MNIST, and the
  external attractors of Section~\ref{subsec:external}). Mean and max of $s_{\max}(W_i)$ over
  all maps, and the percentage of maps above $0.9$ and at or above $1$. Spectral: the
  estimator with spectrally constrained output (Section~\ref{subsec:arch}).
  $^\ddagger$The $100$- and $1000$-step runs of Table~\ref{tab:appendix_training_signal},
  separate from those of Table~\ref{tab:convergence}.}
  \label{tab:contractivity}
  \setlength{\tabcolsep}{3.5pt}
  \begin{tabular}{llrrrrr}
    \toprule
    Setting & Method & Steps & Mean & Max & $>0.9$ (\%) & $\ge1$ (\%) \\
    \midrule
    In distribution & \texttt{base} & 0 & 0.374 & 0.812 & 0.0 & 0.0 \\
     & \texttt{base} & 30 & 0.491 & 0.818 & 0.0 & 0.0 \\
     & \texttt{base} & 100 & 0.547 & 0.904 & 0.1 & 0.0 \\
     & \texttt{base} & 300 & 0.572 & 0.984 & 0.3 & 0.0 \\
     & \texttt{base} & 1000 & 0.576 & 0.964 & 0.3 & 0.0 \\
     & random-r4 & 30 & 0.629 & 0.908 & 0.1 & 0.0 \\
     & random-r1 & 1000 & 0.662 & 0.991 & 0.2 & 0.0 \\
     & random-r4 & 1000 & 0.662 & 0.891 & 0.0 & 0.0 \\
     & spectral ($s_{\max}\le0.95$) & 0 & 0.377 & 0.759 & 0.0 & 0.0 \\
     & spectral ($s_{\max}\le0.95$) & 30 & 0.492 & 0.782 & 0.0 & 0.0 \\
     & \texttt{base}$^\ddagger$ & 100 & 0.544 & 0.900 & 0.1 & 0.0 \\
     & \texttt{base}$^\ddagger$ & 1000 & 0.569 & 1.012 & 0.4 & 0.1 \\
     & no fixed point$^\ddagger$ & 100 & 0.544 & 0.842 & 0.0 & 0.0 \\
     & no fixed point$^\ddagger$ & 1000 & 0.573 & 1.012 & 0.4 & 0.1 \\
     & ordered$^\ddagger$ & 100 & 0.547 & 0.887 & 0.0 & 0.0 \\
     & ordered$^\ddagger$ & 1000 & 0.587 & 1.031 & 0.2 & 0.1 \\
     & multi-hypothesis$^\ddagger$ & 100 & 0.558 & 0.861 & 0.0 & 0.0 \\
     & multi-hypothesis$^\ddagger$ & 1000 & 0.577 & 0.969 & 0.2 & 0.0 \\
     & reconstruction only$^\ddagger$ & 100 & 0.658 & 0.991 & 0.2 & 0.0 \\
     & reconstruction only$^\ddagger$ & 1000 & 0.650 & 1.005 & 0.5 & 0.1 \\
    \midrule
    T1 & \texttt{base} & 0 & 0.353 & 0.712 & 0.0 & 0.0 \\
     & \texttt{base} & 30 & 0.490 & 0.864 & 0.0 & 0.0 \\
     & random-r4 & 60 & 0.672 & 0.926 & 0.4 & 0.0 \\
    \midrule
    S1 & \texttt{base} & 0 & 0.424 & 0.862 & 0.0 & 0.0 \\
     & \texttt{base} & 30 & 0.582 & 0.876 & 0.0 & 0.0 \\
     & random-r4 & 60 & 0.695 & 0.889 & 0.0 & 0.0 \\
    \midrule
    R (reflections) & \texttt{base} & 0 & 0.355 & 0.742 & 0.0 & 0.0 \\
     & \texttt{base} & 30 & 0.485 & 0.819 & 0.0 & 0.0 \\
     & \texttt{base} & 60 & 0.527 & 0.802 & 0.0 & 0.0 \\
     & random-r4 & 60 & 0.657 & 0.899 & 0.0 & 0.0 \\
    \midrule
    P ($\alpha{=}1$) & \texttt{base} & 0 & 0.358 & 0.764 & 0.0 & 0.0 \\
     & \texttt{base} & 30 & 0.473 & 0.787 & 0.0 & 0.0 \\
     & \texttt{base} & 60 & 0.517 & 0.865 & 0.0 & 0.0 \\
     & random-r4 & 60 & 0.639 & 0.917 & 0.1 & 0.0 \\
    \midrule
    MNIST & ours & 0 & 0.311 & 0.579 & 0.0 & 0.0 \\
     & ours & 30 & 0.438 & 0.676 & 0.0 & 0.0 \\
     & ours & 100 & 0.483 & 0.826 & 0.0 & 0.0 \\
     & ours-init $+$ occupancy-GD & 100 & 0.367 & 0.853 & 0.0 & 0.0 \\
    \midrule
    Fashion-MNIST & ours & 0 & 0.352 & 0.726 & 0.0 & 0.0 \\
     & ours & 30 & 0.452 & 0.826 & 0.0 & 0.0 \\
     & ours & 100 & 0.505 & 0.776 & 0.0 & 0.0 \\
     & ours-init $+$ occupancy-GD & 100 & 0.427 & 1.006 & 1.2 & 0.2 \\
    \midrule
    External ($n{=}10$) & ours & 0 & 0.260 & 0.857 & 0.0 & 0.0 \\
     & ours & 100 & 0.462 & 0.768 & 0.0 & 0.0 \\
     & ours & 1000 & 0.520 & 0.772 & 0.0 & 0.0 \\
     & random-r4 & 1000 & 0.582 & 0.776 & 0.0 & 0.0 \\
    \midrule
    Hand-specified ($n{=}4$) & \texttt{base} & 0 & 0.363 & 0.628 & 0.0 & 0.0 \\
     & \texttt{base} & 100 & 0.562 & 0.756 & 0.0 & 0.0 \\
     & random-r4 & 1000 & 0.714 & 0.775 & 0.0 & 0.0 \\
    \bottomrule
  \end{tabular}
\end{table*}

\section{Ambiguity of the inputs}
\label{sec:appendix_ambiguity}
Section~\ref{sec:identifiability} measures non-identifiability between test systems.
To relate it to the inputs of the estimator, we drew a reference set of $131{,}072$
systems from the training generator, disjoint from the training and test data, and
assigned to each \texttt{test256} system with at least three reference systems within
density distance $\varepsilon=0.0168$ (the bottom $1\%$ of test pairs) an ambiguity
score: the median $(W,b)$ set distance to these near-identical inputs. Two caveats
limit what this analysis can show. Only $101$ of the $256$ test systems have such
neighbors, a subset biased toward spread-out, blob-like attractors, and the score is
confounded with this type of image. We therefore report it as a supporting
observation, not as part of the main results.

Two results bear on the framework of the paper (Table~\ref{tab:appendix_ambiguity}).
First, near-identical inputs carry essentially unrelated parameters: the ambiguity
has quartiles $0.95$, $1.03$, and $1.10$, about the mean distance between two unrelated
test systems ($1.06$; Table~\ref{tab:ident_dist}). Second, the $(W,b)$ error of
\texttt{base} does not grow with the ambiguity (Spearman $0.16$ one-shot and $0.04$
after $1000$ steps), whereas the Chamfer distance does ($0.28$ and $0.52$).
If inputs within density distance $\varepsilon=0.0168$ were truly indistinguishable,
a test system and its $\varepsilon$-neighbors would be draws from the same posterior,
and no estimator could reach a mean squared $(W,b)$ error below about $0.54$ on these
$101$ systems (the Bayes risk, half their mean matched squared distance). The one-shot
prediction of \texttt{base} reaches $0.20$ on the same systems, so the density map
carries information about $\theta$ at a scale finer than $\varepsilon$.

\begin{table*}[t]
  \centering\small
  \caption{Ambiguity of the inputs ($101$ \texttt{test256} systems with near-identical
  inputs in the reference set), by quartile of the ambiguity score (1: lowest): mean score, and the
  $(W,b)$ error and Chamfer distance of \texttt{base} one-shot and after $1000$ steps.}
  \label{tab:appendix_ambiguity}
  \setlength{\tabcolsep}{2.5pt}
  \begin{tabular}{crrrrrr}
    \toprule
    Quartile & $n$ & ambiguity & $(W,b)_0$ & $(W,b)_{1000}$ & CD$_0$ & CD$_{1000}$ \\
    \midrule
    1 & 25 & 0.863 & 0.430 & 0.495 & 0.0532 & 0.0168 \\
    2 & 25 & 0.992 & 0.373 & 0.347 & 0.0530 & 0.0169 \\
    3 & 25 & 1.072 & 0.443 & 0.446 & 0.0656 & 0.0218 \\
    4 & 26 & 1.157 & 0.473 & 0.510 & 0.0690 & 0.0250 \\
    \bottomrule
  \end{tabular}
\end{table*}

\section{Point-cloud view of the main result (\texttt{test256})}
\label{sec:appendix_hero_points}
\begin{figure*}[p]
  \centering
  \includegraphics[width=0.6\linewidth]{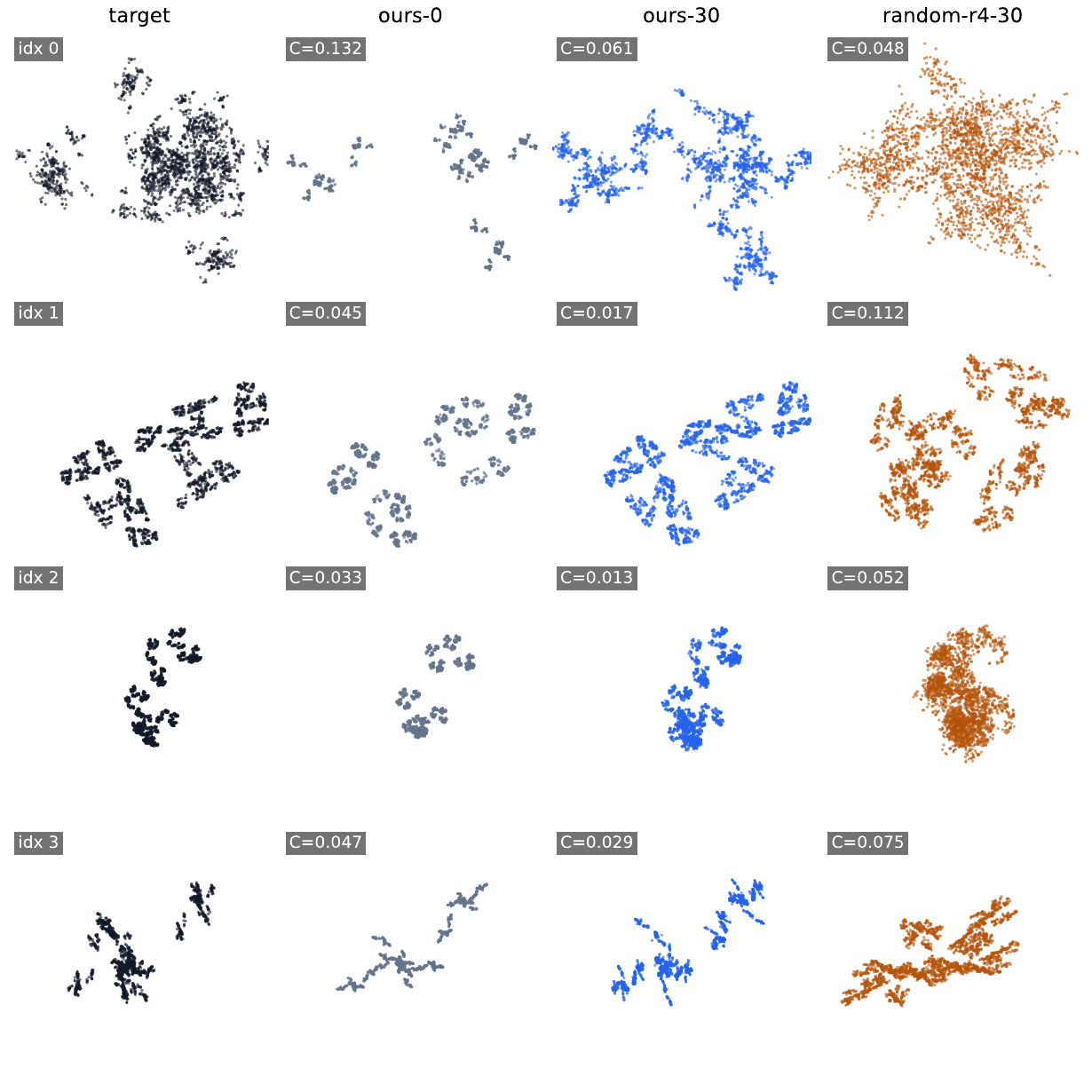}
  \caption{Additional examples of the main result ($4$ examples
  of \texttt{test256}, point-cloud view). Columns: target (black), ours-$0$ (gray),
  ours-$30$ (blue), and random-r4-$30$ (orange); $C$ in each cell is the Chamfer
  distance. These correspond to the same $4$ examples as
  Figure~\ref{fig:hero_qualitative_density}.}
  \label{fig:hero_qualitative_points}
\end{figure*}

We show the same $4$ examples as the qualitative examples of
Section~\ref{subsec:pareto} (Figure~\ref{fig:hero_qualitative_density}, density
maps) as attractor point clouds (Figure~\ref{fig:hero_qualitative_points}). The
point clouds are the trajectory points of the matched renderer (within the drawing
domain $[-1.5,1.5]^2$).

\section{Training signal: all refinement budgets}
\label{sec:appendix_training_signal}
Table~\ref{tab:appendix_training_signal} extends Table~\ref{tab:training_signal} to
every refinement budget and adds the density error and the $(W,b)$ error. All
variants are evaluated with the protocol of Table~\ref{tab:hero_pareto} on the same
\texttt{test256} systems; for the multi-hypothesis head the selected candidate is
refined and scored. The spectrally constrained output (Section~\ref{subsec:arch}) is
included for comparison, and the collage variant is also given with only the
contractivity penalties of reconstruction-only training instead of the output
constraint; both collage variants are taken at the final checkpoint.

\begin{table*}[t]
  \centering
  \small
  \caption{Choice of training signal at all refinement budgets (\texttt{test256},
  matched renderer). Chamfer after $0$--$1000$ steps, density SSE after $30$ steps,
  $(W,b)$ error after $0$ and $30$ steps, and the number of samples (of $256$) on
  which \texttt{base} has the lower Chamfer after $0$, $30$, $100$, and $1000$ steps.}
  \label{tab:appendix_training_signal}
  \setlength{\tabcolsep}{3pt}
  \begin{tabular}{lrrrrrrrrrc}
    \toprule
     & \multicolumn{6}{c}{Chamfer after $k$ steps} & SSE$_{30}$ & \multicolumn{2}{c}{$(W,b)$ error} & \texttt{base} better \\
    \cmidrule(lr){2-7}\cmidrule(lr){9-10}
    Training signal & $0$ & $10$ & $20$ & $30$ & $100$ & $1000$ & ($10^{-4}$) & $0$ & $30$ & $0/30/100/1000$ \\
    \midrule
    \texttt{base}: Hungarian + fixed point & 0.0546 & 0.0365 & 0.0295 & 0.0259 & 0.0201 & 0.0190 & 2.23 & 0.405 & 0.395 & -- \\
    Hungarian, no fixed point & 0.0535 & 0.0365 & 0.0299 & 0.0264 & 0.0202 & 0.0191 & 2.20 & 0.407 & 0.400 & 118 / 135 / 140 / 119 \\
    ordered (fixed-point order) & 0.0852 & 0.0595 & 0.0459 & 0.0384 & 0.0250 & 0.0217 & 3.94 & 0.485 & 0.485 & 211 / 202 / 183 / 155 \\
    multi-hypothesis ($K{=}4$) & 0.0459 & 0.0308 & 0.0255 & 0.0229 & 0.0192 & 0.0192 & 1.72 & 0.429 & 0.427 & 81 / 80 / 111 / 124 \\
    spectrally constrained output & 0.0536 & 0.0357 & 0.0288 & 0.0253 & 0.0196 & -- & 2.08 & 0.403 & 0.397 & 120 / 103 / 121 / -- \\
    \texttt{+aux}: \texttt{base} + reconstruction & 0.0527 & 0.0361 & 0.0297 & 0.0264 & 0.0207 & -- & 2.26 & 0.413 & 0.402 & 94 / 135 / 133 / -- \\
    reconstruction only & 0.0340 & 0.0318 & 0.0311 & 0.0307 & 0.0299 & 0.0295 & 2.65 & 1.072 & 1.062 & 33 / 175 / 217 / 217 \\
    collage only & 0.1011 & 0.0515 & 0.0435 & 0.0411 & 0.0378 & -- & 3.97 & 1.342 & 1.324 & 223 / 225 / 237 / -- \\
    collage only, $s_{\max}\le 0.70$ & 0.0796 & 0.0462 & 0.0412 & 0.0390 & 0.0359 & -- & 3.51 & 1.251 & 1.262 & 193 / 219 / 234 / -- \\
    \bottomrule
  \end{tabular}
\end{table*}

\section{Robustness sweeps}
\label{sec:appendix_robustness}
\begin{figure*}[t]
  \centering
  \includegraphics[width=0.48\linewidth]{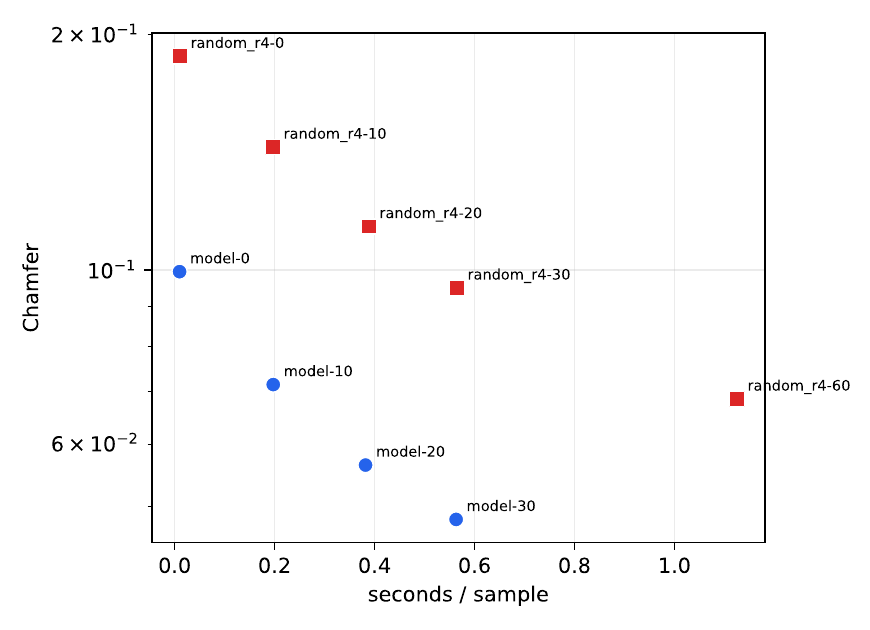}\hfill
  \includegraphics[width=0.48\linewidth]{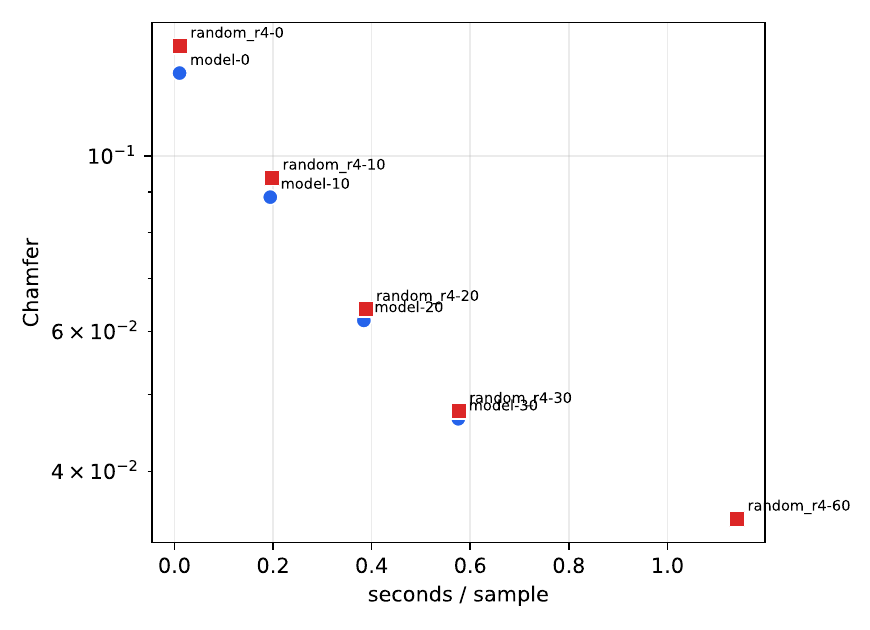}
  \caption{Quality--speed of within-family OOD (Chamfer vs seconds per sample).
  Left: T1 (wide translation); right: S1 (large singular values).}
  \label{fig:ood}
\end{figure*}

Figure~\ref{fig:ood} shows the quality--speed curves of the T1 and S1 shifts of
Table~\ref{tab:ood}.
Table~\ref{tab:appendix_renderer} gives the full observation-renderer sweep
summarized in Table~\ref{tab:renderer}, including refinement with the observation's
renderer settings (renderer known) and observations blurred before inference.
Table~\ref{tab:appendix_prior} gives the reflection and selection-probability shifts
of Table~\ref{tab:ood} for all values of $\alpha$ and with and without the
negative-determinant penalty in refinement. All runs use the \texttt{test256}
systems (R: the training generator with random singular-value signs), $256$ samples,
and the objective, optimizer, and penalties of Section~\ref{subsec:inference}.

\paragraph{Blurring before inference} Blurring a $\sigma{=}1$ observation by
$\sigma{=}2$ gives an effective width of about $2.2$, slightly more than in
training, and restores $0.0306$; blurring it to exactly the training width (by
$\sigma{=}\sqrt3$) restores the in-distribution result ($0.0258$), as does blurring
a $\sigma{=}0$ observation by $\sigma{=}2$.

\paragraph{Reflections} On R, \texttt{base} gives the same result with and without
the negative-determinant penalty in refinement ($0.0318$ either way), so the
penalty does not limit the reconstruction of reflected systems. We also trained
the estimator with the recipe of \texttt{base} on a prior in which each singular
value changes sign with probability $1/2$ (rows ``reflection prior''). It predicts
reflecting maps at about the right rate ($52\%$ of maps, against $55\%$ in R) and has
a smaller $(W,b)$ error ($0.54$ vs $0.67$), but it reconstructs R worse ($0.0797$
one-shot and $0.0397$ after $30$ steps, refined without the negative-determinant
penalty; \texttt{base} better on $213$ and $195$ of $256$ samples). The orientation
of each map adds an ambiguity to the parameter target, and the regression averages
over it: the predicted maps shrink (median total $|\det W|$ $0.24$, against $0.77$
for the true maps), as ordered regression is pulled toward an average in
Section~\ref{subsec:training_signal}.

\begin{table*}[t]
  \centering\small
  \caption{Observation-renderer sweep (\texttt{test256}). One setting of the
  observation renderer is changed at a time from the training renderer ($16$
  trajectories of $1024$ steps, burn-in $128$, hard histogram, $\sigma{=}2$, $128^2$).
  Chamfer of ours at $0$ and $30$ steps, HD95 and coverage@2px of ours-30, Chamfer of
  random-r4-30, and the number of samples (of $256$) on which ours-30 has the lower
  Chamfer. Unless stated, refinement uses the training renderer.}
  \label{tab:appendix_renderer}
  \begin{tabular}{lrrrrrr}
    \toprule
    Observation & ours-0 & ours-30 & HD95$_{30}$ & cov$_{30}$ & r4-30 & wins \\
    \midrule
    training renderer & 0.0546 & 0.0258 & 0.0920 & 0.860 & 0.0726 & 250 \\
    $2$ trajectories & 0.0575 & 0.0273 & 0.0970 & 0.849 & 0.0728 & 254 \\
    $4$ trajectories & 0.0567 & 0.0269 & 0.0960 & 0.852 & 0.0728 & 251 \\
    $8$ trajectories & 0.0549 & 0.0260 & 0.0917 & 0.861 & 0.0728 & 252 \\
    $64$ trajectories & 0.0543 & 0.0259 & 0.0920 & 0.858 & 0.0731 & 248 \\
    $256$ trajectories & 0.0540 & 0.0257 & 0.0914 & 0.862 & 0.0730 & 250 \\
    $256{\times}4096$ points & 0.0541 & 0.0258 & 0.0918 & 0.861 & 0.0730 & 250 \\
    resolution $64^2$ & 0.0600 & 0.0305 & 0.1128 & 0.828 & 0.0734 & 248 \\
    resolution $96^2$ & 0.0556 & 0.0272 & 0.0978 & 0.851 & 0.0729 & 251 \\
    resolution $256^2$ & 0.0544 & 0.0256 & 0.0907 & 0.864 & 0.0726 & 250 \\
    bilinear splatting & 0.0546 & 0.0258 & 0.0916 & 0.863 & 0.0726 & 249 \\
    no burn-in & 0.0546 & 0.0257 & 0.0912 & 0.863 & 0.0730 & 250 \\
    $\sigma{=}0$ & 0.2032 & 0.1166 & 0.4678 & 0.483 & 0.0716 & 61 \\
    $\sigma{=}1$ & 0.1456 & 0.0756 & 0.2933 & 0.588 & 0.0721 & 135 \\
    $\sigma{=}3$ & 0.0708 & 0.0379 & 0.1480 & 0.787 & 0.0745 & 239 \\
    $\sigma{=}4$ & 0.0728 & 0.0423 & 0.1709 & 0.771 & 0.0773 & 239 \\
    $\sigma{=}1$, pre-blurred by $2$ & 0.0604 & 0.0306 & 0.1136 & 0.825 & 0.0730 & 250 \\
    $\sigma{=}1$, pre-blurred by $\sqrt3$ & 0.0545 & 0.0258 & 0.0919 & 0.861 & 0.0730 & 251 \\
    $\sigma{=}0$, pre-blurred by $2$ & 0.0546 & 0.0258 & 0.0920 & 0.861 & 0.0726 & 250 \\
    $2$ trajectories, refined with obs.\ renderer & 0.0575 & 0.0273 & 0.0968 & 0.847 & 0.0726 & 255 \\
    $4$ trajectories, refined with obs.\ renderer & 0.0567 & 0.0271 & 0.0969 & 0.850 & 0.0731 & 251 \\
    $\sigma{=}0$, refined with obs.\ renderer & 0.2032 & 0.1281 & 0.5121 & 0.465 & 0.0758 & 46 \\
    $\sigma{=}4$, refined with obs.\ renderer & 0.0728 & 0.0395 & 0.1572 & 0.781 & 0.0761 & 242 \\
    \bottomrule
  \end{tabular}
\end{table*}

\begin{table*}[t]
  \centering\small
  \caption{Prior shifts outside the training distribution (Chamfer, $256$ samples
  each). R: reflections; P: selection probabilities $p=(1-\alpha)\,p_{\det}+\alpha\,q$
  with $q\sim\mathrm{Dirichlet}(1,\dots,1)$. The last column counts the samples on
  which ours-30 beats random-r4-30. In distribution ($\alpha{=}0$): ours-0 $0.0546$,
  ours-30 $0.0258$, random-r4-30 $0.0726$. Reflection prior: the estimator trained
  with reflections in its prior (see text); on \texttt{test256},
  which has none, it gives $0.0820$ and $0.0448$ at $0$ and $30$ steps.}
  \label{tab:appendix_prior}
  \begin{tabular}{lrrrrrr}
    \toprule
    Shift & ours-0 & ours-30 & ours-60 & r4-30 & r4-60 & wins \\
    \midrule
    reflections (det.\ penalty $0.1$) & 0.0606 & 0.0318 & 0.0270 & 0.0758 & 0.0533 & 251 \\
    reflections (det.\ penalty $0$) & 0.0606 & 0.0318 & 0.0270 & 0.0758 & 0.0533 & 250 \\
    reflections, reflection prior (det.\ penalty $0$) & 0.0797 & 0.0397 & 0.0301 & 0.0758 & 0.0533 & 241 \\
    reflections, reflection prior (det.\ penalty $0.1$) & 0.0797 & 0.0469 & 0.0437 & 0.0759 & 0.0532 & 218 \\
    probabilities, $\alpha{=}0.25$ & 0.0610 & 0.0295 & 0.0245 & 0.0732 & 0.0512 & 252 \\
    probabilities, $\alpha{=}0.5$ & 0.0619 & 0.0312 & 0.0267 & 0.0738 & 0.0516 & 252 \\
    probabilities, $\alpha{=}0.75$ & 0.0630 & 0.0331 & 0.0291 & 0.0749 & 0.0531 & 244 \\
    probabilities, $\alpha{=}1$ & 0.0632 & 0.0354 & 0.0326 & 0.0789 & 0.0577 & 248 \\
    \bottomrule
  \end{tabular}
\end{table*}

\section{External IFS: details}
\label{sec:appendix_external}
\paragraph{Protocol} The renderer of Djeacoumar et al.\ normalizes each attractor to
its bounding box, so their codes are defined only up to a similarity. We therefore
conjugate the ground truth and each published code by the similarity that fits its
attractor to the same box in our domain and draw them with the benchmark's
probability rule ($\propto|\det W|+0.01$). Our estimator receives the observation of
the normalized ground truth, and its outputs, like those of random-r4, are drawn with
our own rule (Eq.~\eqref{eq:prob_det}); all results are scored with our point-cloud
metrics.

\begin{figure*}[t]
  \centering
  \includegraphics[width=0.78\linewidth]{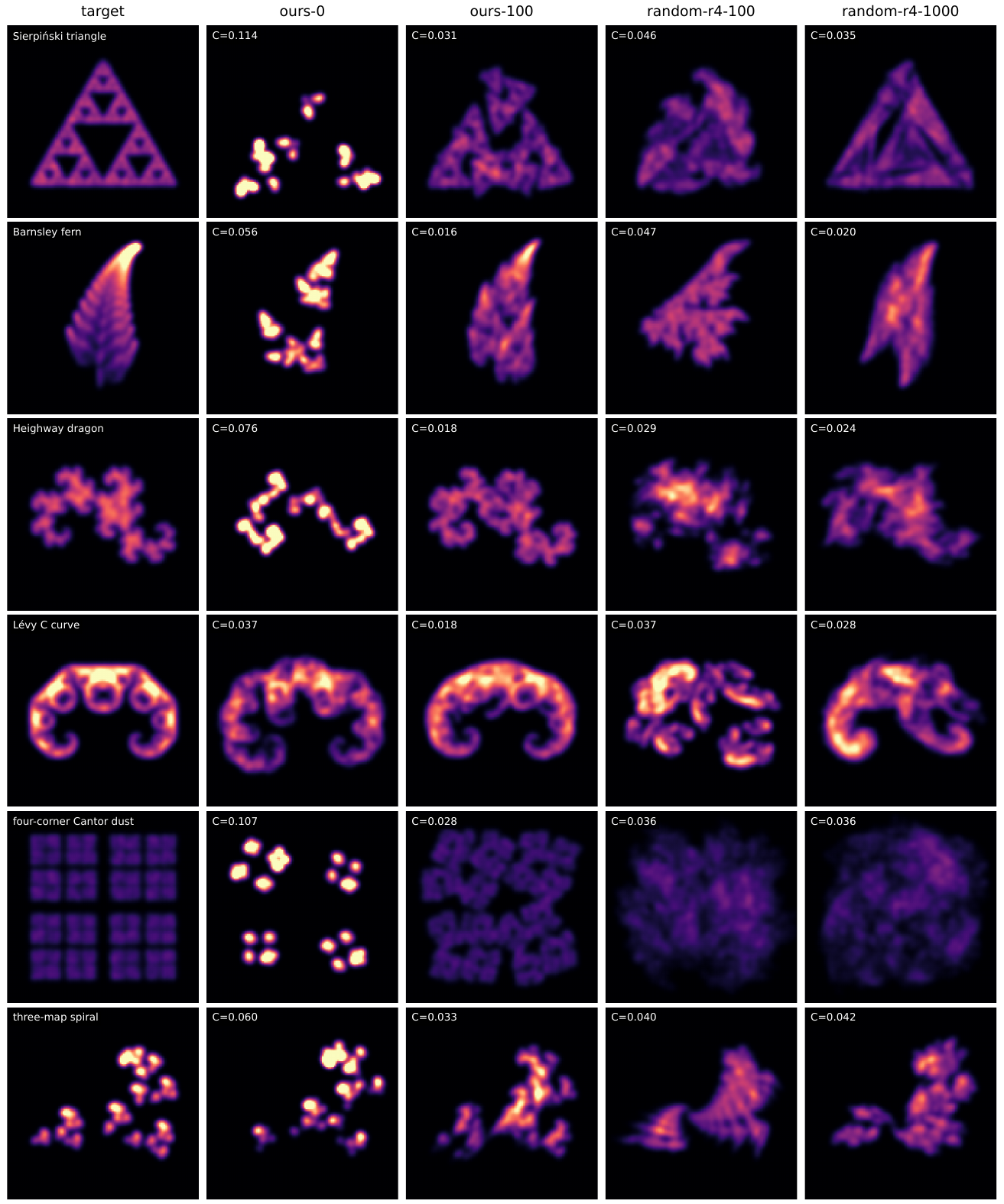}
  \caption{Hand-specified IFS: targets drawn with their standard probabilities (the
  spiral with $p\propto|\det W|+0.01$) and
  the predictions of our $n{=}4$ model ($0$ and $100$ refinement steps) and of
  random-r4 ($100$ and $1000$ steps), all on the standard renderer. C: Chamfer to the
  true attractor.}
  \label{fig:appendix_classical}
\end{figure*}

\paragraph{Benchmark and hand-specified IFS} Table~\ref{tab:appendix_external_n4} evaluates the $43$ IFS of the Djeacoumar et al.\
benchmark with at most four maps under our $n{=}4$ model (the other methods keep their
$10$-map codes, which favors them). The picture is the same as for all $101$ IFS with
the $n{=}10$ model (Table~\ref{tab:external}): refinement from the amortized prediction
beats random starts at every budget, the mean after $1000$ steps is on par with
Djeacoumar et al., and their optimization is better on most individual IFS.
The six hand-specified IFS are four classical fractals (the Sierpi\'nski triangle,
the Barnsley fern, the Heighway dragon, and the L\'evy C curve) and two simple
constructions (a four-corner Cantor dust and a three-map spiral), drawn with their
standard probabilities (for the spiral, which we constructed,
$p\propto|\det W|+0.01$). Table~\ref{tab:appendix_classical} and
Figure~\ref{fig:appendix_classical} give them individually. The Barnsley fern is the hardest case for the
prior: one of its maps is singular, one reverses orientation, one has $s_{\max}=0.85$,
and its published probabilities differ from $|\det W|$. The Heighway dragon and the
L\'evy C curve have $s_{\max}=0.71$, just above the training range.

\paragraph{Alignment check} The codes published by Djeacoumar et al.\ are aligned with
the truth through the bounding boxes of their attractors (see Protocol above),
whereas our predictions are made on the aligned observation, so errors of the box
estimate penalize only their codes. As a post-hoc check, we therefore applied to each
attractor produced by their method and by ours (after $100$ and $1000$ steps) the
uniform scale and translation that minimize its Chamfer distance to the truth, fitted
numerically ($300$ Adam steps) on one sampled point cloud and evaluated on an
independent one. On the newly drawn evaluation point clouds, this alignment changes
the mean Chamfer of Djeacoumar et al.\ from $0.0209$ to $0.0198$, the median from
$0.0178$ to $0.0175$, and the number of IFS above $0.03$ from $14$ to $10$; ours after
$1000$ steps changes from $0.0200$ to $0.0196$ (median $0.0199$ to $0.0193$, worst
$0.031$ to $0.030$), and ours after $100$ steps from $0.0209$ to $0.0205$. Against
ours after $1000$ steps, their optimization is better on $66$ of the $101$ IFS before
alignment and $69$ after. The values before alignment differ slightly from
Table~\ref{tab:external} because the point clouds are redrawn; the per-IFS count is
the most sensitive to this ($66$ here against $62$ in Table~\ref{tab:external}).

\begin{table*}[t]
  \centering\small
  \caption{Benchmark IFS with at most four maps ($43$ IFS), evaluated with our $n{=}4$
  model. Columns as in Table~\ref{tab:external}.}
  \label{tab:appendix_external_n4}
  \begin{tabular}{lrrrrrrr}
    \toprule
    Method & CD mean & CD median & HD95 & cov@2px & success$_{1\,\mathrm{px}}$ & better & non-contr.\ (\%) \\
    \midrule
    ours, $0$ steps    & 0.0828 & 0.0802 & 0.3206 & 0.583 & 0.00 & 1  & 0 \\
    ours, $30$ steps   & 0.0361 & 0.0333 & 0.1302 & 0.790 & 0.12 & 5  & 0 \\
    ours, $100$ steps  & 0.0255 & 0.0246 & 0.0918 & 0.877 & 0.44 & 9  & 0 \\
    ours, $1000$ steps & \textbf{0.0224} & 0.0216 & \textbf{0.0784} & 0.906 & 0.63 & 16 & 0 \\
    random-r4, $100$ steps  & 0.0334 & 0.0313 & 0.1274 & 0.837 & 0.05 & 3  & 0 \\
    random-r4, $1000$ steps & 0.0248 & 0.0244 & 0.0904 & 0.885 & 0.42 & 11 & 0 \\
    \midrule
    Djeacoumar et al.  & 0.0227 & \textbf{0.0189} & 0.0827 & \textbf{0.940} & \textbf{0.74} & -- & 0 \\
    evolutionary       & 0.0941 & 0.0424 & 0.3676 & 0.716 & 0.02 & 0 & 49 \\
    cuckoo search      & 0.0848 & 0.0756 & 0.3322 & 0.633 & 0.00 & 1 & 74 \\
    neural regression  & 0.1314 & 0.1259 & 0.4863 & 0.440 & 0.00 & 0 & 5 \\
    \bottomrule
  \end{tabular}
\end{table*}

\begin{table*}[t]
  \centering\small
  \caption{Hand-specified IFS (Chamfer; $n{=}4$ model). Floor: the true IFS
  re-rendered.}
  \label{tab:appendix_classical}
  \setlength{\tabcolsep}{3pt}
  \begin{tabular}{lrrrrr}
    \toprule
    IFS & floor & ours-0 & ours-100 & r4-100 & r4-1000 \\
    \midrule
    Sierpi\'nski & 0.0079 & 0.1141 & \textbf{0.0308} & 0.0455 & 0.0349 \\
    fern & 0.0094 & 0.0557 & \textbf{0.0162} & 0.0474 & 0.0199 \\
    dragon & 0.0110 & 0.0757 & \textbf{0.0182} & 0.0292 & 0.0245 \\
    L\'evy C & 0.0107 & 0.0370 & \textbf{0.0183} & 0.0374 & 0.0282 \\
    Cantor dust & 0.0160 & 0.1065 & \textbf{0.0283} & 0.0363 & 0.0355 \\
    spiral & 0.0040 & 0.0597 & \textbf{0.0332} & 0.0401 & 0.0420 \\
    \midrule
    mean & 0.0099 & 0.0748 & \textbf{0.0242} & 0.0393 & 0.0308 \\
    \bottomrule
  \end{tabular}
\end{table*}

\paragraph{Official code on \texttt{test256}} For Section~\ref{subsec:external} we ran
the official implementation of Djeacoumar et al.\ (unchanged apart from a one-line
build fix) with its default configuration ($10$ maps, $15{,}000$ iterations, seed $5$)
on the first $16$ \texttt{test256} systems (Table~\ref{tab:appendix_i7}). Each target
is drawn by their ground-truth generator ($5\times10^{7}$ points on a $1024^2$ grid with
$16\times$ supersampling, probabilities $\propto|\det W|+0.01$, normalized to the
bounding box), the optimized code is mapped back to our frame through the bounding box
of its attractor as for the benchmark, and all methods are scored against the same
ground-truth point clouds. Their optimization took $2993$\,s per system on average on an
otherwise idle RTX~3090. Scoring their codes against ground truth drawn with their
probability rule changes their mean Chamfer by less than $0.001$. With the best-fit
scale and translation of the alignment check above (and new point samples), their
mean becomes $0.0475$ (median $0.0340$), ours after $30$ and $1000$ steps $0.0271$ and
$0.0205$, and ours remains better on $9$ and $13$ of the $16$ systems. One of their
runs collapsed to ten near-point maps ($s_{\max}=0.07$, Chamfer $0.25$); without it,
their mean is $0.041$.

\begin{figure*}[t]
  \centering
  \includegraphics[width=\linewidth]{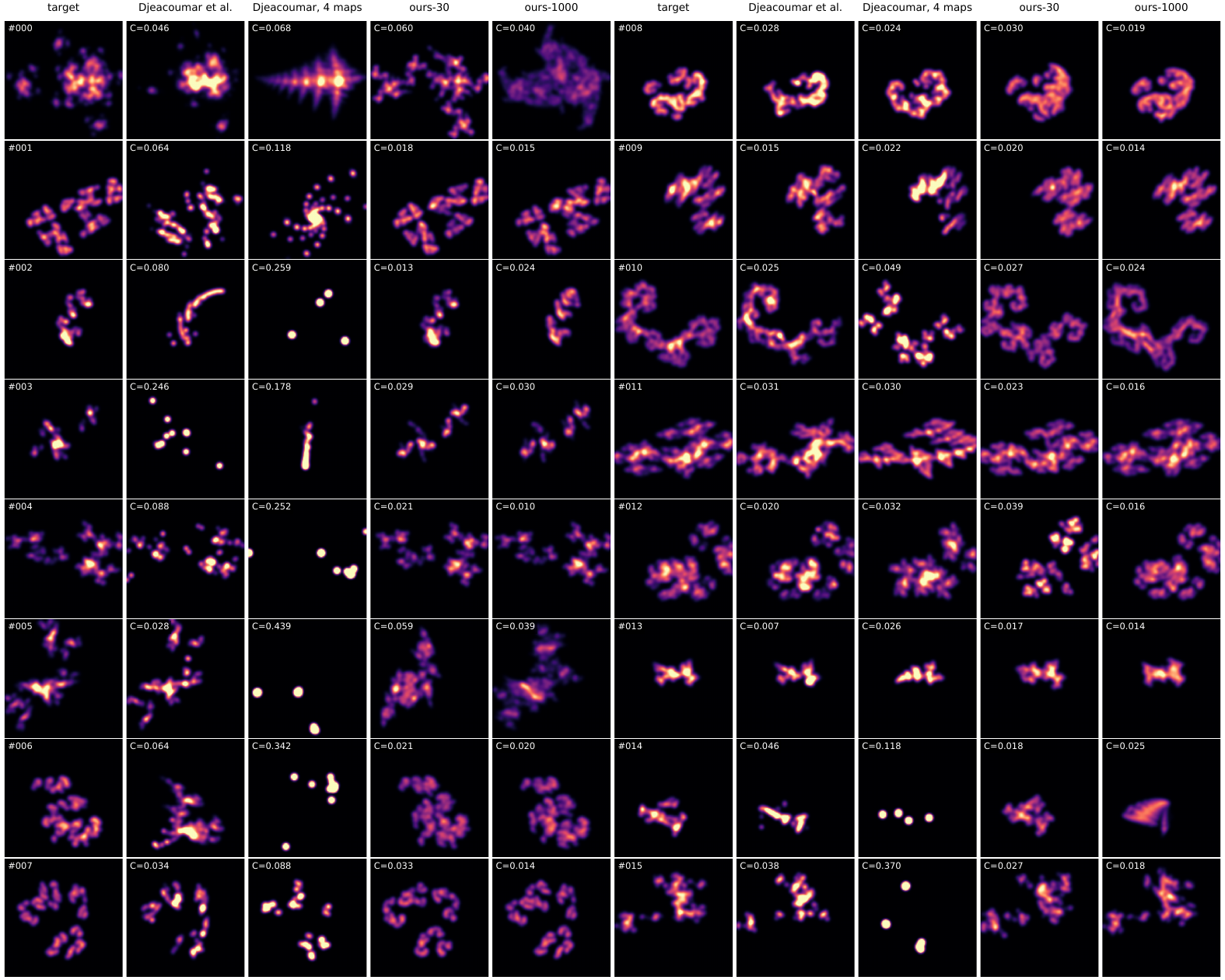}
  \caption{Official code of Djeacoumar et al.\ on the $16$ \texttt{test256} systems (two
  blocks of eight): target, their result with the default configuration ($10$ maps) and
  with the true number of maps ($4$), and ours after $30$ and $1000$ steps, all drawn on
  our standard renderer in our frame. C: Chamfer to the true attractor.}
  \label{fig:appendix_i7}
\end{figure*}

\paragraph{Protocol checks} Two checks address whether this result reflects the
systems or our use of their code. First, we reran their optimizer on the same $16$
systems with the true number of maps ($4$) instead of the default $10$. It does
worse (Table~\ref{tab:appendix_i7}): $13$ of the $16$ results lie above Chamfer $0.03$,
four runs collapse to near-point maps, and only $3$ systems improve on the $10$-map
run, so the over-complete default helps their method. Second, we ran their code with
its default configuration on the six most strongly contracting systems of their own
benchmark ($s_{\max}$ from $0.50$ to $0.66$), on their own target images, and scored
our runs and the codes they publish for these systems in the same way. Our runs are
worse on all six (mean Chamfer $0.0333$ vs $0.0277$, median $0.0302$ vs $0.0242$), so
the published results may come from a more favorable setting than the default one we
use, and our protocol may understate their method by about this margin, which is much
smaller than the gap on \texttt{test256} ($0.0537$ vs $0.0211$). The systems
themselves are the more likely cause. All \texttt{test256} systems contract strongly
($s_{\max}\le0.70$), and on their benchmark the published results fail most often on
the systems with $s_{\max}<0.75$ ($6$ of $17$ above $0.03$, against $9$ of the other
$84$); within the $16$ \texttt{test256} systems their error grows as the attractor
becomes sparser (Spearman $-0.69$ with $\sum_i|\det W_i|$). Figure~\ref{fig:appendix_i7}
shows their outputs, which often reproduce only the densest parts of these sparse
attractors or concentrate the mass in a few compact clusters.

\begin{table*}[t]
  \centering\small
  \caption{Official code of Djeacoumar et al.\ on the first $16$ \texttt{test256}
  systems ($n{=}4$, inside our training prior). Columns as in
  Table~\ref{tab:external}; better: systems on which the method has a lower Chamfer
  than Djeacoumar et al.\ with the default configuration ($10$ maps). Floor: the true
  IFS re-rendered.}
  \label{tab:appendix_i7}
  \begin{tabular}{lrrrrrr}
    \toprule
    Method & CD mean & CD median & HD95 & cov@2px & success$_{1\,\mathrm{px}}$ & better \\
    \midrule
    ours, $0$ steps    & 0.0659 & 0.0503 & 0.2624 & 0.628 & 0.00 & 6 \\
    ours, $30$ steps   & 0.0283 & 0.0250 & 0.1011 & 0.845 & 0.50 & 9 \\
    ours, $100$ steps  & 0.0220 & 0.0204 & 0.0792 & 0.894 & \textbf{0.75} & 10 \\
    ours, $1000$ steps & \textbf{0.0211} & \textbf{0.0186} & \textbf{0.0760} & \textbf{0.909} & 0.63 & 14 \\
    random-r4, $1000$ steps & 0.0303 & 0.0290 & 0.1215 & 0.842 & 0.25 & 11 \\
    Djeacoumar et al.  & 0.0537 & 0.0359 & 0.1944 & 0.773 & 0.19 & -- \\
    Djeacoumar et al., $4$ maps & 0.1509 & 0.1031 & 0.4802 & 0.516 & 0.06 & -- \\
    \midrule
    floor (re-rendered) & 0.0075 & 0.0071 & 0.0208 & 0.996 & 1.00 & -- \\
    \bottomrule
  \end{tabular}
\end{table*}

\section{Additional examples (MNIST / Fashion-MNIST)}
\label{sec:appendix_qualitative}
\begin{figure*}[p]
  \centering
  \includegraphics[height=0.84\textheight]{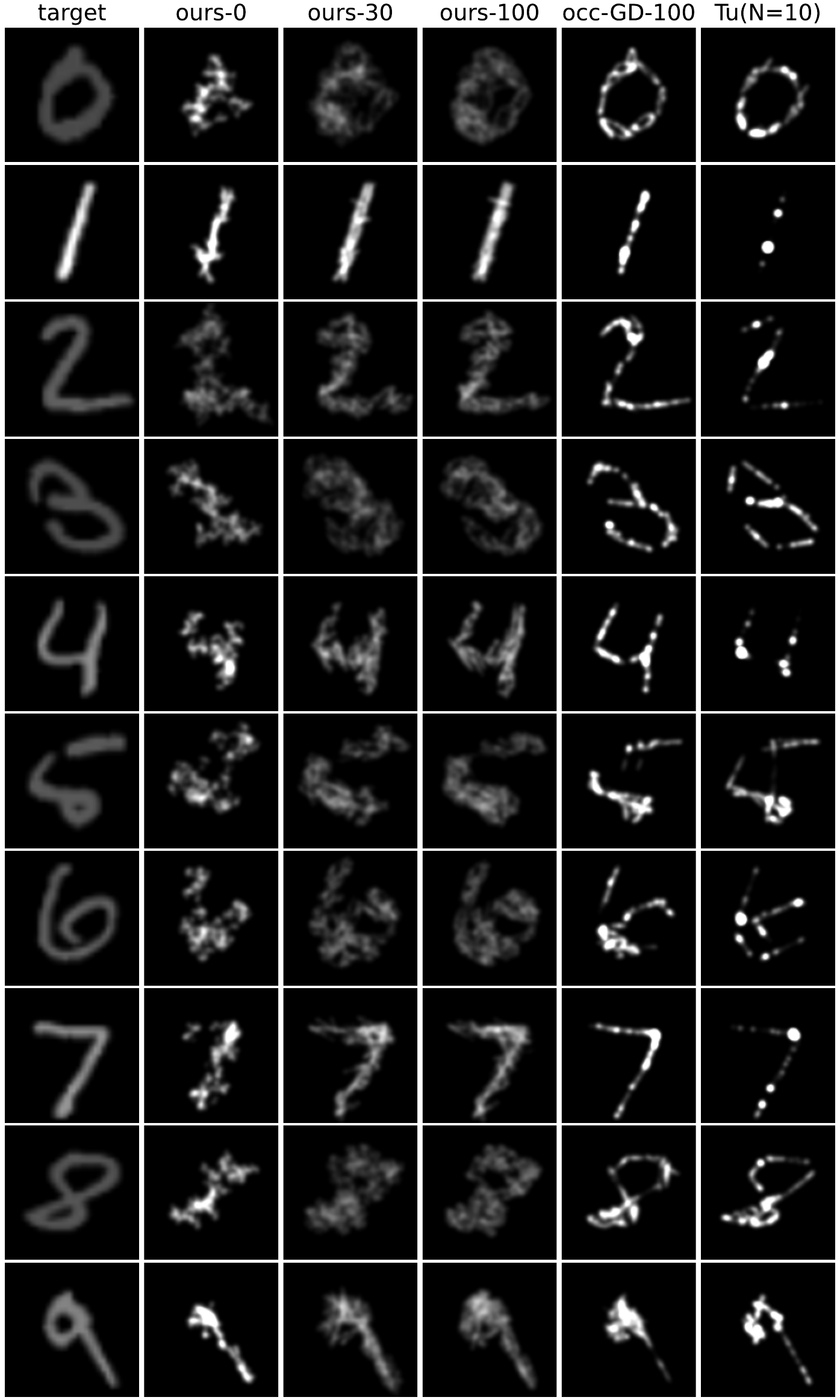}
  \caption{MNIST reconstruction (density maps, common grayscale, one example per
  row). Columns from left: target, ours $0$-step, ours $30$-step, ours $100$-step,
  ours-init$+$occupancy-GD $100$, Tu ($n{=}10$). Under density-aware refinement,
  mass spreads as an area and approaches the target's thick density as
  $0\to30\to100$ steps, whereas Tu and occupancy-GD have clear outlines but
  concentrate mass on thin lines and isolated points.}
  \label{fig:mnist_density}
\end{figure*}

\begin{figure*}[p]
  \centering
  \includegraphics[height=0.84\textheight]{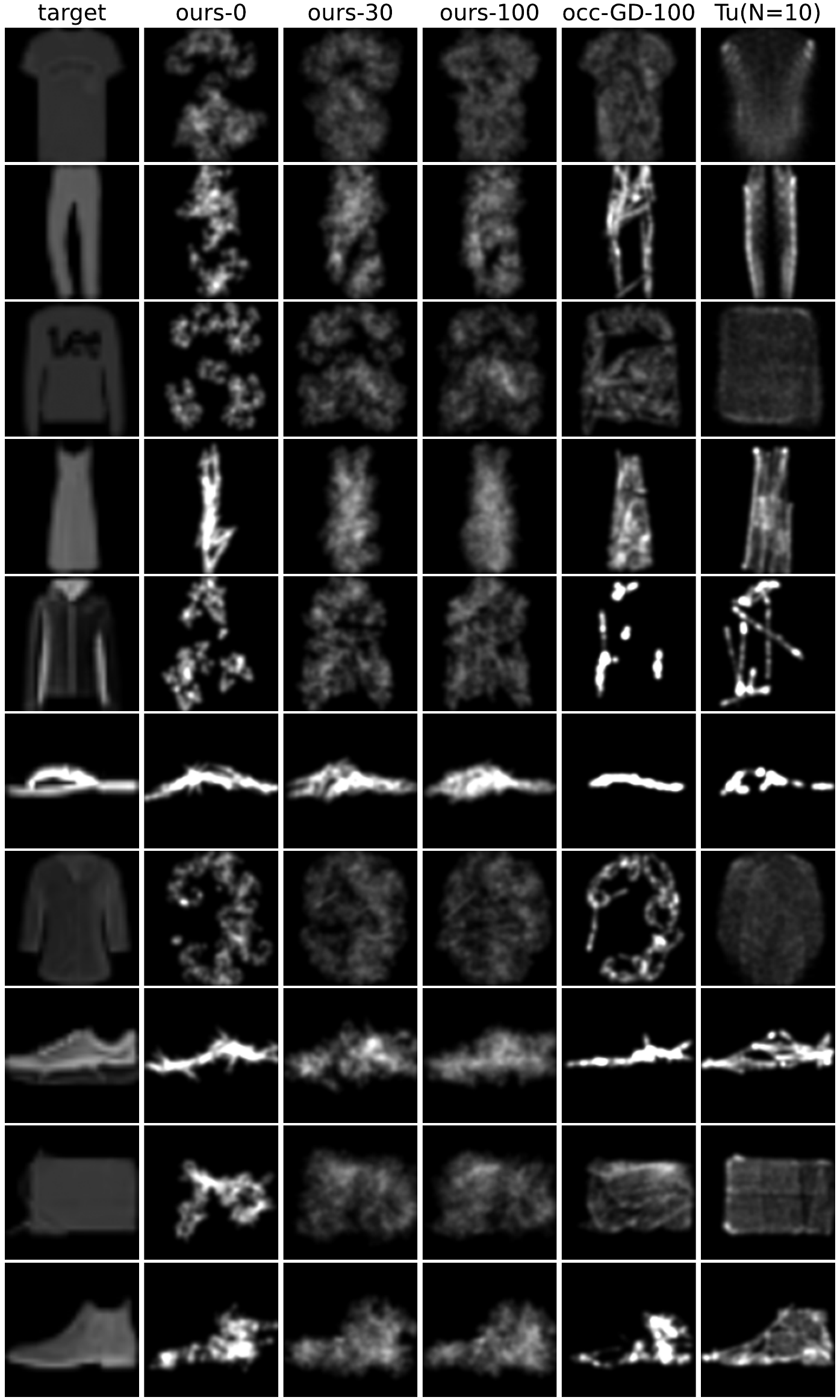}
  \caption{Fashion-MNIST reconstruction (density maps, common grayscale, one example
  per row). Columns from left: target, ours $0$-step, ours $30$-step, ours
  $100$-step, ours-init$+$occupancy-GD $100$, Tu ($n{=}10$). Faithful reconstruction
  of filled regions requires reproducing internal density; density refinement
  ($0\to30\to100$) reproduces the target's area density well, whereas occupancy-GD
  damages density and collapses the garment to an outline.}
  \label{fig:fmnist_density}
\end{figure*}

\begin{figure*}[p]
  \centering
  \includegraphics[height=0.84\textheight]{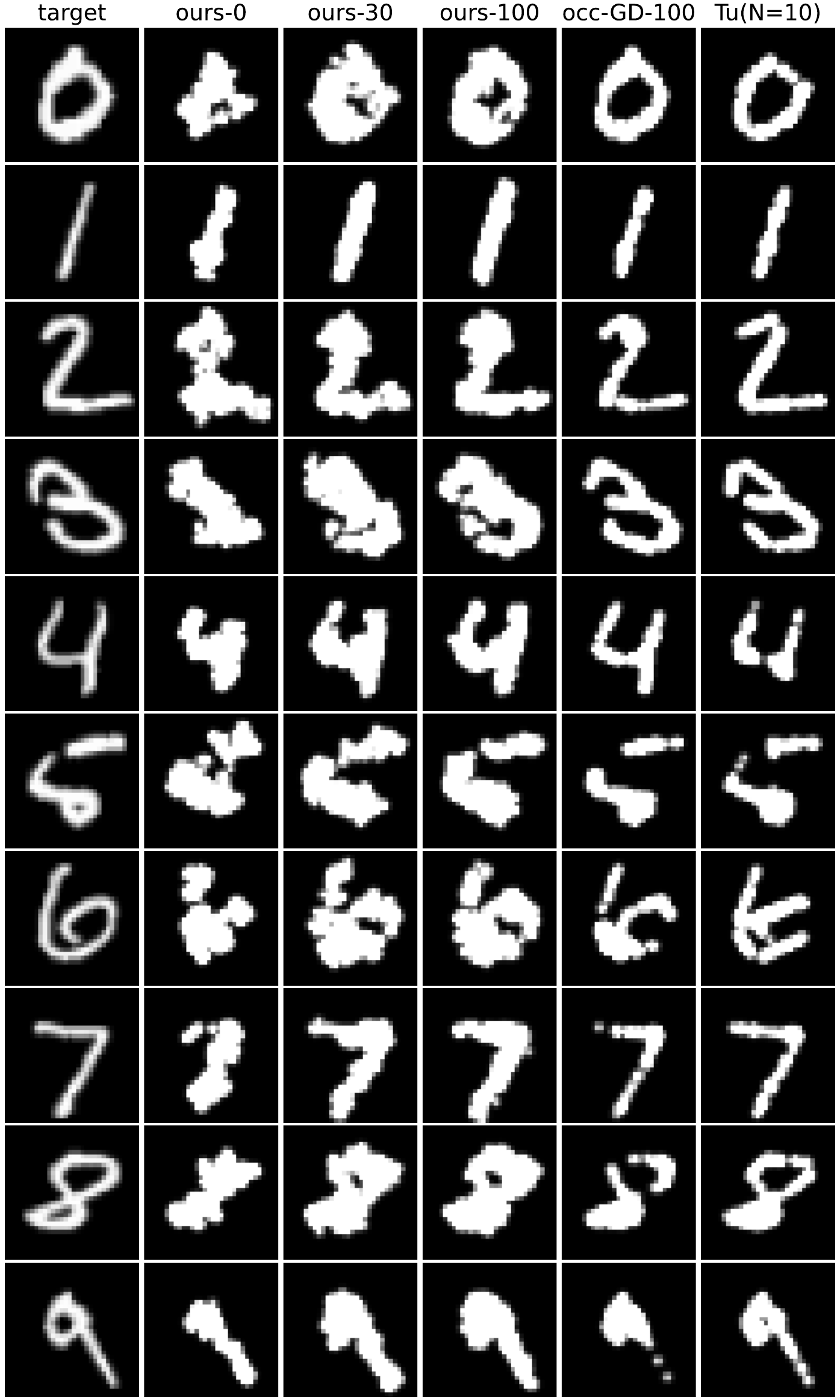}
  \caption{MNIST, rendered with Tu's renderer ($32\times32$ saturated occupancy)
  ($10$ examples). Columns are the same as
  Figure~\ref{fig:mnist_density} (target / ours-0 / ours-30 / ours-100 / occupancy-GD-100
  / Tu ($n{=}10$)).}
  \label{fig:appendix_occupancy_mnist}
\end{figure*}

\begin{figure*}[p]
  \centering
  \includegraphics[height=0.84\textheight]{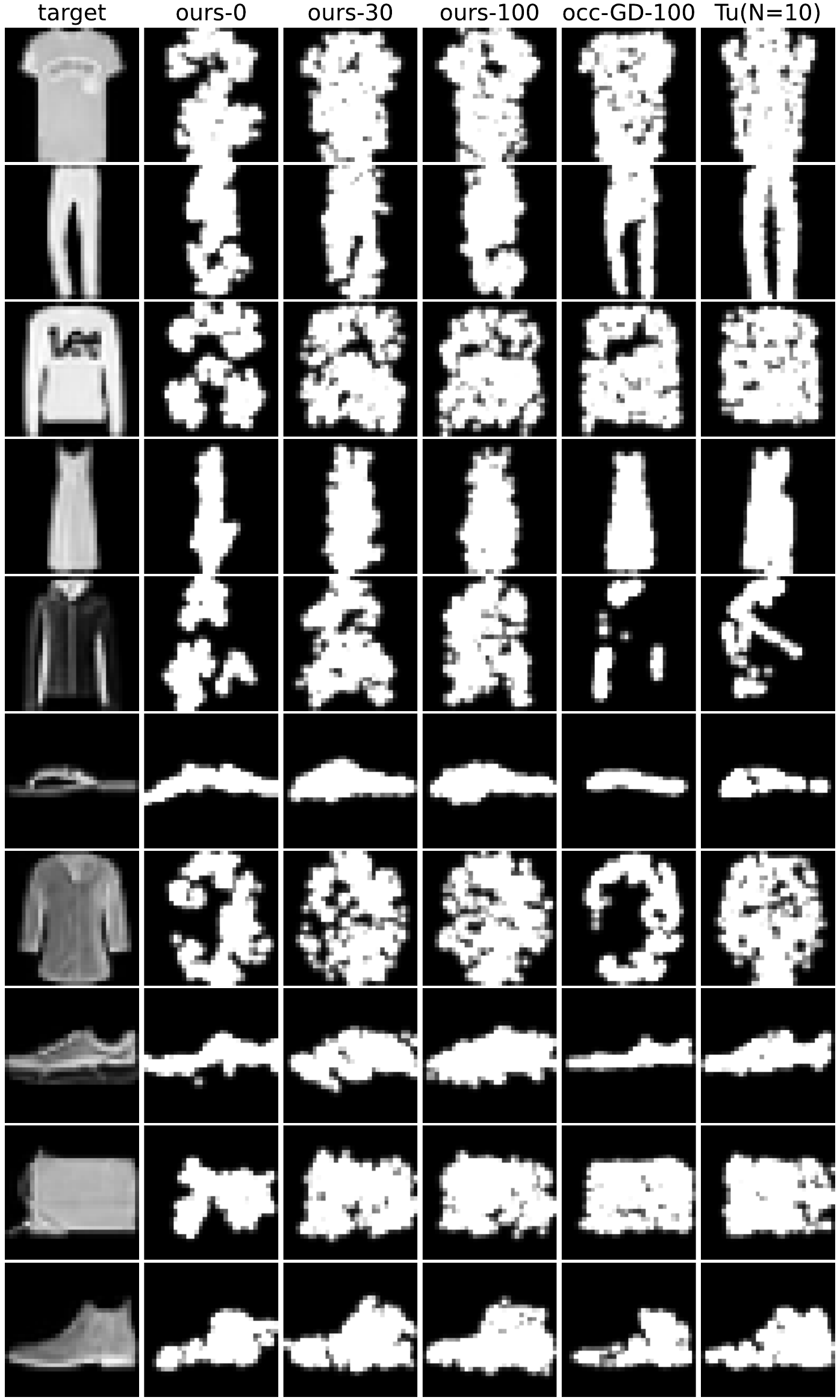}
  \caption{Fashion-MNIST, rendered with Tu's renderer ($32\times32$ saturated
  occupancy) ($10$ examples). Columns are the same as
  Figure~\ref{fig:fmnist_density}.}
  \label{fig:appendix_occupancy_fmnist}
\end{figure*}

\begin{figure*}[p]
  \centering
  \includegraphics[height=0.84\textheight]{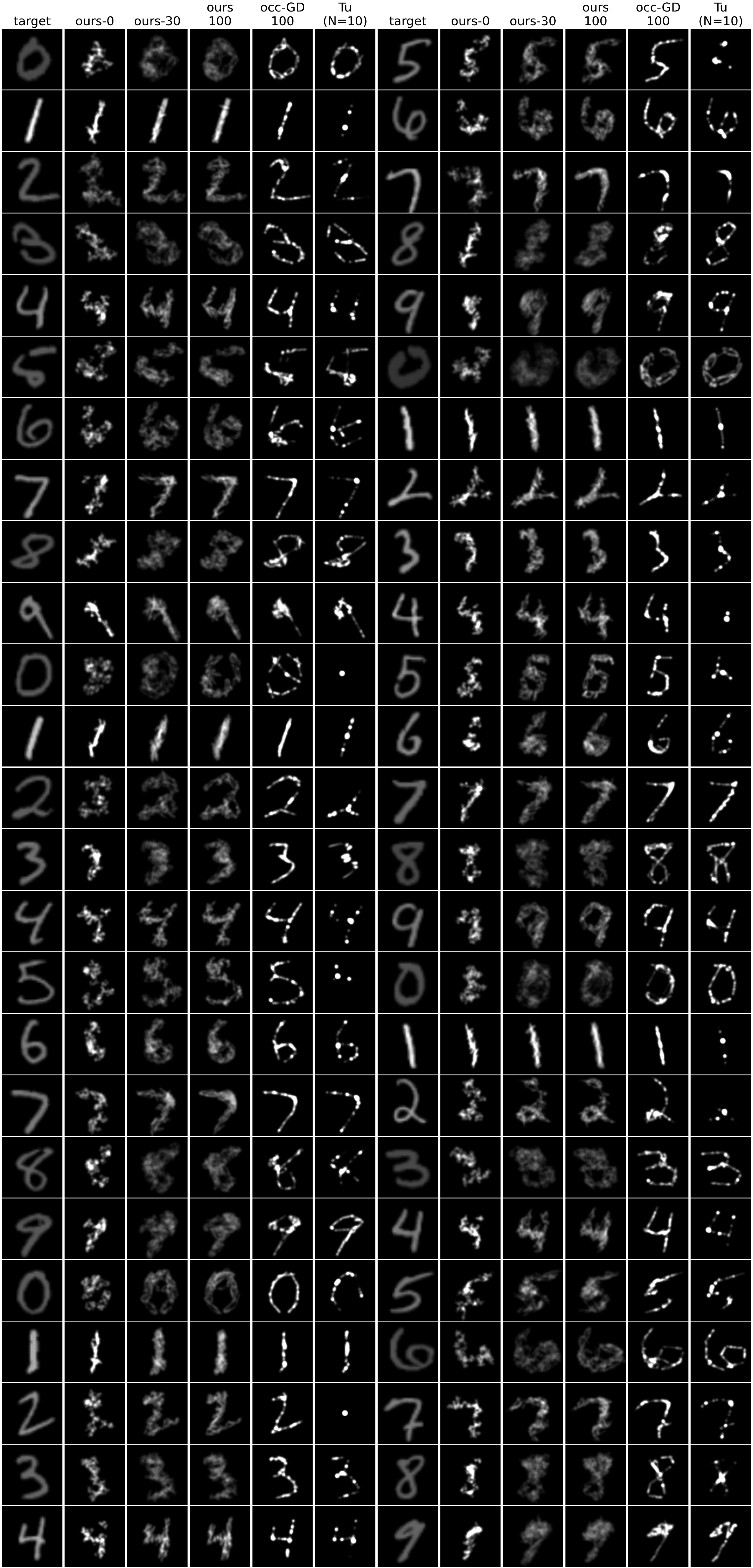}
  \caption{MNIST, density maps (common grayscale), all balanced $50$ examples.
  Columns are the same as Figure~\ref{fig:mnist_density}.}
  \label{fig:appendix_density50_mnist}
\end{figure*}

\begin{figure*}[p]
  \centering
  \includegraphics[height=0.84\textheight]{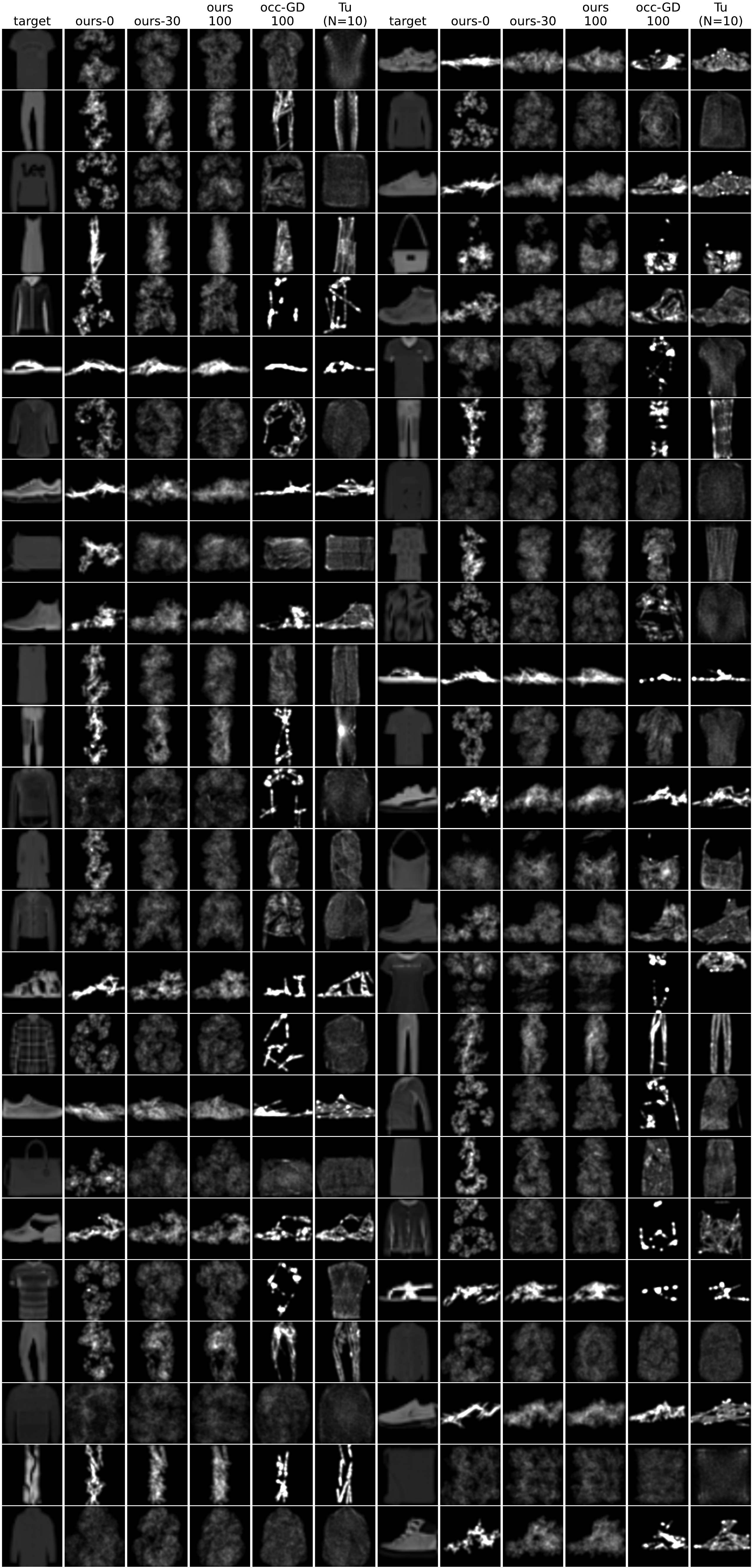}
  \caption{Fashion-MNIST, density maps (common grayscale), all balanced $50$
  examples. Columns are the same as Figure~\ref{fig:fmnist_density}.}
  \label{fig:appendix_density50_fmnist}
\end{figure*}

For the comparison of Section~\ref{subsec:mnist_tu}, we show additional qualitative
examples. Figures~\ref{fig:mnist_density} and~\ref{fig:fmnist_density} give the ten
examples in the density-map view whose first five are in Figure~\ref{fig:real_density}.
We add (i) an occupancy view of the same outputs rendered with Tu's renderer
($32\times32$ saturated occupancy)
(Figures~\ref{fig:appendix_occupancy_mnist} and~\ref{fig:appendix_occupancy_fmnist}),
and (ii) density galleries of all balanced $50$ examples
(Figures~\ref{fig:appendix_density50_mnist} and~\ref{fig:appendix_density50_fmnist}).
In the occupancy view, Tu often shapes outlines clearly, but there are also examples
where it drops thin lines or concentrates mass excessively onto lines and points. In
the density view, Tu's outputs often have the clearest outlines, while ours-$100$
better preserves intermediate tones in some examples, and the tendency of our
refinement ($0\to30\to100$) to reproduce the target's area density well is
consistent across the $50$ examples.

\clearpage
\end{document}